%% file: main.tex
\documentclass[sigconf]{acmart}
\usepackage[normalem]{ulem}
\usepackage{multirow}
\usepackage{enumitem}
\usepackage{amsthm}

\theoremstyle{definition} 

\usepackage{tikz}
\usepackage{booktabs}
\usepackage{multirow}
\usepackage{graphicx}
\usepackage[table]{xcolor}
\usepackage{fontawesome5} 
\usepackage{array}
\usepackage{pifont}
\usepackage{longtable}
\usepackage{tcolorbox}
\usepackage{listings}
\usepackage{xcolor}
\usepackage{tabularx}   
\usepackage{ragged2e}   
\newcommand{\gHigh}{\cellcolor{black!25}} 
\newcommand{\gGood}{\cellcolor{black!18}} 
\newcommand{\gMid}{\cellcolor{black!10}}  
\newcommand{\gLow}{\cellcolor{black!4}}   
\newcommand{\gBad}{\cellcolor{white}}     

\usepackage{wasysym}      

\newcolumntype{Y}{>{\centering\arraybackslash}p{2.2cm}}

\newcolumntype{L}{>{\scriptsize}l}

\newcommand{\fullcirc}{\CIRCLE}
\newcommand{\halfcirc}{\LEFTcircle}
\newcommand{\emptycirc}{\Circle}

\AtBeginDocument{%
  }

\setcopyright{none}
\renewcommand\footnotetextcopyrightpermission[1]{}
\acmConference[]{}{}{}
\begin{document}

\newcommand{\ModelName}{\textsc{SciHorizon-DataEVA}}
\newcommand{\ModelPrinciplesName}{Sci-TQA² principles}
\newcommand{\ModelEvalName}{Sci-TQA²-Eval}
\newcommand{\degreecircle}[1]{%
  \tikz\draw[fill=black!#1,draw=black] (0,0) circle (0.12);%
}



\title{\ModelName: An Agentic System to Scalable AI-Readiness Evaluation of Heterogeneous Scientific Data}


\author{Dianyu Liu, Chuan Qin, Xi Chen, Xiaohan Li, \\Wenxi Xu, Yuyang Wang, Xin Chen, Yuanchun Zhou, Hengshu Zhu*}
\thanks{Corresponding author.}
\affiliation{%
 \institution{SciHorizon Team, Computer Network Information Center, Chinese Academy of Sciences}
 \city{Beijing}
 \country{China}
 }

\renewcommand{\shortauthors}{Liu et al.}

\input{0_abstract}

\maketitle
\pagestyle{plain}
\input{1_Introduction}

\input{2_RelatedWork}
\input{4_Sci-principles}

\input{5_Sci-Eval}
\input{6_Experiments}
\input{7_Conclusion}

\bibliographystyle{ACM-Reference-Format}
\balance
\bibliography{reference}
\newpage
\input{8_Appendix}

\end{document}

%% file: 0_abstract.tex
\begin{abstract}
    AI-for-Science (AI4Science) is increasingly transforming scientific discovery by embedding machine learning models into prediction, simulation, and hypothesis generation workflows across domains. However, the effectiveness of these models is fundamentally constrained by the AI-readiness of scientific data, for which no scalable and systematic evaluation mechanism currently exists. In this work, we propose \ModelName, a novel agentic system to scalable AI-readiness evaluation of heterogeneous scientific data. At the evaluation-criteria level, we introduce the \ModelPrinciplesName, which organize AI-readiness into four complementary dimensions: Governance Trustworthiness, Data Quality, AI Compatibility, and Scientific Adaptability. Each dimension is decomposed into measurable atomic elements that enable fine-grained and executable assessment. To operationalize these principles at scale, we develop \ModelEvalName, a hierarchical multi-agent evaluation approach orchestrated through a directed, cyclic workflow. Our \ModelEvalName dynamically constructs dataset-aware evaluation specifications by combining lightweight dataset profiling, applicability-aware metric activation, and knowledge-augmented planning grounded in domain constraints and dataset–paper signals. These specifications are executed through an adaptive, tool-centric evaluation mechanism with built-in verification and self-correction, enabling scalable and reliable assessment across heterogeneous scientific data. Extensive experiments on scientific datasets spanning multiple domains demonstrate the effectiveness and generality of \ModelName\ for principled AI-readiness evaluation.
\end{abstract}

%% file: 1_Introduction.tex
\section{Introduction}

AI-for-Science (AI4Science) has emerged as a core research paradigm in which machine learning models are embedded within scientific analysis and discovery workflows~\cite{wei2025agentic, zheng2025automation, chen2025ai4research,long2026sciencedb}. Across domains such as structural biology and materials science, AI methods are increasingly used for prediction, simulation, and data-driven hypothesis generation~\cite{jumper2021highly,merchant2023scaling}. The effectiveness of these methods, however, is fundamentally constrained by the AI-readiness of the scientific data they operate on. Scientific datasets vary widely in quality, governance, and structural characteristics, and such heterogeneity directly affects their suitability for AI training, inference, and downstream scientific reasoning~\cite{bodnar2025datareadiness}. Despite the rapid growth of large-scale scientific data repositories—such as ScienceDB, which hosts over 15 million datasets as of January 2026~\cite{ScienceDB2026}—there is no scalable mechanism to assess AI-readiness, leaving data selection largely ad hoc and limiting effective use at scale.

Data evaluation has long been studied in the data mining literature. Classical data quality frameworks primarily assess generic properties such as completeness, consistency, and timeliness~\cite{wang1996beyond}, while FAIR-oriented tools focus on findability, accessibility, interoperability, and reusability of data~\cite{wilkinson2016fair}. 
Recently, SciHorizon framework~\cite{scihorizonbenchmarkingaiforscience2025qin} introduced a novel ``Data + Large Language Model'' dual-perspective approach to assess AI4Science readiness, including an AI-readiness assessment framework for scientific data. However, the current evaluation process still relies heavily on manual participation and lacks an automated, intelligent workflow for scalable assessment. Moreover, the indicator system remains insufficient for comprehensively evaluating AI-readiness in scientific contexts, and the framework primarily focuses on Earth and Life Sciences, limiting its extensibility across diverse scientific domains. Consequently, existing approaches exhibit fundamental limitations in \textbf{what they evaluate}. \textsl{One key limitation is the lack of assessment of AI compatibility.} Current methods do not assess whether data representations, feature structures, and implicit inductive assumptions are compatible with the requirements of learning and inference. As a result, datasets that satisfy traditional quality criteria may still be unsuitable for modern AI models~\cite{chen2026beyond,tong2025missteps,shen2026prompting,qin2025cotr,jiang2024enhancing,chen-etal-2026-gendis,xu-etal-2026-tlsa,qin-etal-2026-bolt,zhu2025survey}. For example, many molecular property predictors rely on 3D conformations and equivariant representations, rendering datasets that provide only SMILES strings or 2D graphs effectively incompatible despite being structurally complete~\cite{liu2023interpretable, long2024moat}.
\textsl{Equally important, existing approaches rarely evaluate scientific adaptability.} They do not determine whether datasets contain the contextual and experimental variables required for scientifically meaningful reasoning about underlying mechanisms, rather than mere statistical fitting. Consequently, models trained on such data may produce predictions that appear accurate yet fail to support scientific interpretation or generalization. In clinical datasets, for instance, missing socioeconomic variables can lead models to attribute patient outcomes to hospital choice rather than underlying social or clinical factors, yielding predictions that are statistically plausible but scientifically invalid~\cite{prosperi2020causal}. \looseness=-1

\input{tables/intro}

Beyond these limitations in existing evaluation criteria, current tools face substantial practical challenges in \textbf{how} the AI-readiness of scientific data \textbf{is evaluated}. \textsl{A central difficulty stems from the heterogeneous and multimodal nature of scientific data.} Scientific datasets span diverse modalities, including genomic sequences, medical images, molecular graphs, and high-dimensional simulation outputs, each governed by distinct structural and semantic constraints. As a result, most existing evaluation tools are designed for a narrow set of data types and fail to generalize across modalities without extensive manual adaptation. \textsl{These challenges are further exacerbated by limited scalability.} Existing evaluation mechanisms are predominantly static, relying on manually specified rules or rigid scripts that are typically designed for specific data formats or structured representations. Applying such tools to heterogeneous scientific datasets often requires domain experts to develop specialized evaluation logic for each new data type or discipline. As scientific data continues to grow in both volume and diversity, this expert-driven, format-specific process becomes increasingly costly and difficult to sustain at scale.

To address the above challenges, we propose a novel agentic system, called \textbf{\ModelName}, for scalable AI-readiness evaluation of heterogeneous scientific data.
At the evaluation-criteria level, we introduce \textbf{\ModelPrinciplesName}, which organizes AI-readiness into four complementary dimensions: \textsl{Governance Trustworthiness} (T), \textsl{Data Quality} (Q), \textsl{AI~Compatibility} (A), and \textsl{Scientific Adaptability}~(A). Each dimension is further decomposed into measurable atomic elements that serve as executable units for systematic evaluation. Building on this foundation, we develop \textbf{\ModelEvalName}, a hierarchical multi-agent evaluation approach that operationalizes these principles at scale through a directed, cyclic workflow. Specifically, \textbf{\ModelEvalName} first generates dataset-aware evaluation specifications: a profile-oriented data inspector constructs a lightweight dataset profile without full data loading; an applicability-aware metric selector activates only fine-grained evaluation metrics that are feasible for the given dataset; and a knowledge-augmented evaluation specification planner instantiates each activated atomic element into executable data loading and scoring strategies, guided by retrieved domain constraints (Type-I knowledge) and dataset–paper relational signals (Type-II knowledge). These specifications are then executed by an adaptive, tool-centric evaluation module that integrates a structured tool library, persistent tool memory, and extensible tool orchestration to support scalable execution and cold-start tool generation, while a review-driven verification and self-correction mechanism ensures runtime correctness and semantic validity. Finally, element-level results are aggregated into multi-dimensional AI-readiness scores and synthesized into an AI-Ready Report, enabling robust and generalizable readiness assessment across heterogeneous scientific data. Extensive experiments on scientific datasets spanning multiple scientific domains demonstrate the effectiveness of our approach.

%% file: tables/intro.tex
\begin{table*}[t]
\centering

\renewcommand{\arraystretch}{0.9} 

\setlength{\tabcolsep}{2pt}
\vspace{2mm}
\caption{Comparison of \textbf{\ModelName} with existing data evaluation frameworks.}
\label{tab:comparison}

\resizebox{\textwidth}{!}{%
\begin{tabular}{L|YYYYYYY} 
\toprule

\textbf{\normalsize Methods} & 
\textbf{\small DQ Toolkit} & 
\textbf{\small FAIR-Cookbook} & 
\textbf{\small FAIR-Checker} & 
\textbf{\small OpenDataVal} & 
\textbf{\small AIDRIN} & 
\textbf{\small SciHorizon} & 
\textbf{\small Ours} \\

& 
\scriptsize{\cite{dqlearntoolkitstructured2020shrivastavab}} & 
\scriptsize{\cite{faircookbookessential2023rocca-serra}} & 
\scriptsize{\cite{faircheckersupportingdigital2023gaignard}} & 
\scriptsize{\cite{opendatavalunifiedbenchmark2023jiang}} & 
\scriptsize{\cite{aidatareadiness2024hiniduma}} & 
\scriptsize{\cite{scihorizonbenchmarkingaiforscience2025qin}} & 
\scriptsize{ } \\

\midrule

\multicolumn{8}{l}{\textit{\textbf{\small Dimensions}}} \\ 
\quad 1. Trustworthiness & \halfcirc & \fullcirc & \fullcirc & \emptycirc & \fullcirc & \fullcirc & \fullcirc \\
\quad 2. General Quality & \fullcirc & \emptycirc & \halfcirc & \fullcirc & \fullcirc & \fullcirc & \fullcirc \\
\quad 3. AI Compatibility & \halfcirc & \emptycirc & \emptycirc & \fullcirc & \fullcirc & \halfcirc & \fullcirc \\
\quad 4. Scientific Adaptability & \emptycirc & \emptycirc & \emptycirc & \emptycirc & \emptycirc & \halfcirc & \fullcirc \\

\midrule

\multicolumn{8}{l}{\textit{\textbf{\small Capabilities}}} \\
\quad Multi-Modal Support & \emptycirc & \emptycirc & \halfcirc & \halfcirc & \emptycirc & \fullcirc & \fullcirc \\
\quad Automated Tools & \fullcirc & \emptycirc & \fullcirc & \fullcirc & \fullcirc & \halfcirc & \fullcirc \\
\quad Adaptability & \emptycirc & \halfcirc & \halfcirc & \emptycirc & \emptycirc & \emptycirc & \fullcirc \\

\bottomrule
\end{tabular}%
}
\\[2ex]
\footnotesize{
\textit{Symbols}: \fullcirc~Fully Supported, \halfcirc~Partially Supported, \emptycirc~Unsupported.
}
\end{table*}

%% file: 2_RelatedWork.tex
\section{Related Work}

\subsection{Scientific Dataset Evaluation}
Data evaluation is a cornerstone of building reliable data-driven systems. Early work largely framed it as a classical data quality problem, focusing on intrinsic dimensions such as accuracy, completeness, and consistency~\cite{intrinsicdataquality2025haug, overviewimportancedata2020jaina}. In parallel, the FAIR principles (Findable, Accessible, Interoperable, Reusable) became the de facto standard for scientific data stewardship~\cite{designframeworkexemplar2018wilkinson, faircookbookessential2023rocca-serra}. These paradigms have substantially shaped how datasets are curated and shared.

Recently, however, a growing body of work has argued that traditional quality notions and FAIRness, while necessary, are not sufficient for data used in AI model training and deployment. Beyond being “high quality” or “FAIR,” datasets must satisfy AI-specific requirements related to robustness, bias, generalization, and downstream utility~\cite{zhou2025enhancing,zhu2025enhancing,qin2025survey,fang2023recruitpro,long2026survey,fang2021spatial,zhang2026fairgc, zhu2025can}. This has led to the emergence of \emph{AI-Readiness} as a distinct evaluation paradigm that explicitly targets the suitability of data for AI pipelines~\cite{datareadinessai2025hiniduma}.
Several domain-specific AI-Readiness frameworks have been proposed in areas such as space science, medicine, and broader scientific data management. These frameworks introduce readiness levels, trustworthiness dimensions, and lifecycle-oriented maturity matrices~\cite{aireadydataspace2023poduval, metricframeworkassessingdata2024schwabe, datareadinessscientific2025brewer}. Most notably, the SciHorizon project established a benchmark across Earth, Life, and Material sciences~\cite{scihorizonbenchmarkingaiforscience2025qin}.
However, these approaches typically cover only a subset of relevant dimensions, rely heavily on predefined tools with limited coverage, and often require substantial expert intervention. In this paper, we propose holistic evaluation principles, \ModelPrinciplesName, which explicitly address this gap.

\subsection{Automated Evaluation Methods}

Existing evaluation practices rely heavily on predefined tools, which can be grouped into three categories: traditional data quality tools, scientific metadata evaluators, and static AI-readiness toolkits.
General tools such as DQLearn~\cite{dqlearntoolkitstructured2020shrivastavab} and the Data Quality Toolkit~\cite{dataqualitytoolkit2021gupta} provide APIs for profiling structured data and detecting statistical anomalies (e.g., label noise), but lack the domain context required for scientific data.
For scientific data management, tools such as FAIR-Checker~\cite{faircheckersupportingdigital2023gaignard} use semantic web technologies to audit metadata against community standards (e.g., DCAT, Bioschemas). They are effective for assessing FAIR compliance but do not inspect data content or its suitability for AI modeling.
More recent tools such as AIDRIN~\cite{aidatareadiness2024hiniduma, aidrin20framework2025hiniduma} and OpenDataVal~\cite{opendatavalunifiedbenchmark2023jiang} integrate AI-specific metrics (e.g., fairness, privacy, feature importance) into unified frameworks. However, they still rely on static, predefined rules, are mainly designed for structured data, and struggle to assess heterogeneous scientific modalities (e.g., molecular graphs, genomic sequences) or to generate domain-specific evaluation tools autonomously.

Recent advances in multi-agent systems (MAS) have demonstrated strong reasoning capabilities and sophisticated tool-use in scientific domains~\cite{virtuallabai2025swanson, honeycombflexiblellmbased2024zhang, sciagenttoolaugmentedlanguage2024ma, maladeorchestrationllmpowered2024choi, luo2025large}. Building on these, we design \ModelEvalName, which is, to our knowledge, the first system to apply agentic tool generation to adaptively assess AI-readiness—including AI compatibility across heterogeneous scientific datasets.

%% file: 4_Sci-principles.tex
\section{Sci-TQA² Principles}
\label{sec:tqa2}
We propose \ModelPrinciplesName, which systematically characterizes the AI-readiness of scientific data into four dimensions: \textsl{Governance Trustworthiness}, \textsl{Data Quality}, \textsl{AI Compatibility}, and \textsl{Scientific Adaptability}, each further decomposed into sub-dimensions and atomic evaluation elements.


\subsection{Top-level Dimensions}

\subsubsection{Governance Trustworthiness}
This dimension characterizes whether a scientific data can be safely reused, shared, and redistributed under legal, ethical, and institutional constraints. It focuses on governance artifacts that render the dataset lifecycle transparent and auditable, which is particularly critical in high-stakes settings (e.g., clinical data or proprietary industrial experiments), where non-compliant reuse may invalidate results or incur legal risk.


This dimension is specified through three sub-dimensions: \textsl{FAIRness Principles}, which address the findability, accessibility, interoperability, and reusability of data and metadata; \textsl{Provenance \& Licensing}, which ensure traceability of data origin and clarity of ownership and usage rights; and \textsl{Scientific Ethics}, which cover consent compliance and dual-use risk considerations. At a practical level, Governance Trustworthiness evaluates whether downstream users can unambiguously determine \emph{what the data is}, \emph{where it originates}, \emph{under what terms it may be used}, and \emph{whether its use is ethically permissible}. Strong performance on this dimension lowers non-technical barriers to adoption and supports cross-lab reproducibility through explicit and verifiable governance.


\subsubsection{Data Quality}
This dimension assesses whether a dataset is technically reliable and sufficiently standardized for ingestion by computational pipelines with minimal manual intervention. It reflects how faithfully the released dataset represent the intended observations and whether they conform to domain-appropriate representations, such as unit systems, coordinate conventions, and identifier standards. Data quality has a direct impact on downstream engineering effort, as deficiencies often necessitate substantial reconciliation across files, instruments, or modalities.

This dimension comprises four sub-dimensions: \textsl{Completeness}, which addresses missing values and record integrity; \textsl{Accuracy}, which concerns the correctness of measurements and annotations; \textsl{Uniqueness}, which captures the presence of duplicate or near-duplicate samples; and \textsl{Consistency}, which ensures uniform formats, schemas, and encodings across artifacts. By systematically evaluating these sub-dimensions, practitioners can identify concrete quality deficiencies and prioritize targeted remediation actions, such as data collection or imputation, calibration and validation, deduplication, and unit or schema normalization, enabling curation-oriented improvements beyond coarse, one-shot quality indicators.



\subsubsection{AI Compatibility}
This dimension assesses the technical utility of a dataset for supporting learning and inference in AI models. Beyond basic data cleanliness, a dataset must provide sufficient learning signal, appropriate representational structure, and adequate scale to support common modeling paradigms. This dimension therefore examines whether a dataset can be effectively used by AI systems under realistic modeling assumptions.

We further decompose it into four sub-dimensions: \textsl{AI Model Adaptability}, which reflects support for diverse model families and learning setups; \textsl{Data Scale}, which captures effective sample size and dimensionality; \textsl{Class Balance}, which measures skew in label or outcome distributions; and \textsl{Feature Importance}, which assesses the strength of association between inputs and targets.
By evaluating these sub-dimensions, practitioners can derive modeling-oriented recommendations, including adopting data-efficient learning when scale is limited, applying reweighting or resampling under severe class imbalance, and reconsidering modeling objectives when features are weakly informative. Such evaluation also helps identify bottlenecks that render standard learning and inference unreliable.



\subsubsection{Scientific Adaptability}
This dimension evaluates whether a dataset supports genuine scientific discovery rather than mere statistical fitting. Whereas AI Compatibility concerns the feasibility of learning and inference, Scientific Adaptability examines whether a dataset enables mechanism-aware reasoning, meaningful generalization across regimes, and scientifically interpretable conclusions.

This dimension is decomposed into four sub-dimensions. \textsl{Task Generalizability} captures the breadth of downstream scientific tasks supported by the dataset. \textsl{Data Scarcity} reflects the difficulty of data acquisition and the uniqueness of observations. \textsl{Causal Completeness} evaluates whether key variables and potential confounders are sufficiently represented to support causal reasoning. \textsl{Operating Condition Coverage} assesses the extent to which the dataset spans relevant physical or experimental parameter ranges. A dataset may be large and well annotated yet still score poorly on this dimension if it omits critical regimes, conflates confounding factors, or lacks the contextual variables required to interpret outcomes. By evaluating these sub-dimensions, Scientific Adaptability estimates a dataset’s potential to support scientific discovery and signals when a benchmark risks encouraging narrow interpolation rather than robust scientific insight.



\subsection{Atomic Evaluation Elements}
\label{sec:atomic-elements}

To support fine-grained, actionable assessment, each sub-dimension in \ModelPrinciplesName~is decomposed into atomic evaluation elements. These elements represent the smallest measurable units of evaluation and, when feasible, are implemented as standalone quantitative checks. Some atomic elements are domain-agnostic (e.g., file integrity, schema validity, basic missingness rates, duplicate detection, and train--test leakage checks), whereas others require domain expertise for meaningful specification and quantification (e.g., acceptable value ranges, instrument-specific calibration constraints, and scientifically meaningful coverage of operating regimes).

This atomic-element design ensures that evaluation remains both \emph{mathematically well-defined} and \emph{scientifically grounded}. Each element produces a normalized score accompanied by interpretable metrics and evidence traces (e.g., the fraction of records that violate unit conventions, the fields that lack provenance, and the subsets that exhibit distribution shift). Element-level scores can be aggregated to obtain sub-dimension and dimension-level summaries while preserving drill-down diagnostics for remediation. 


%% file: 5_Sci-Eval.tex
\section{\ModelEvalName}
To operationalize the above principles for scientific data evaluation in an automated and scalable manner across heterogeneous datasets, we design a novel evaluation approach, \ModelEvalName. 
As shown in Figure \ref{fig:framework}, it is implemented as an MAS orchestrated by a directed cyclic graph (DCG) workflow. Unlike static, single-pass pipelines, \ModelEvalName~tightly integrates \emph{knowledge-grounded planning}, \emph{tool-centric execution}, and \emph{review-driven feedback}, enabling robust and adaptive evaluation over diverse scientific data representations and domain contexts. We first give an overview of the system architecture, followed by detailed descriptions of each module.

\begin{figure*}
    \centering
    \includegraphics[width=\linewidth]{figs/framework.pdf}
    \caption{The architecture overview of our \ModelEvalName.}
    \label{fig:framework}
\end{figure*}
\subsection{Overview}
Our MAS consists of two coupled modules, transforming a raw dataset into executable quality assessments and an AI-ready report:
\paragraph{(I) Dataset-Aware Evaluation Specification Generation with Knowledge Augmentation.}
We first build a data profile $\mathcal{P}_{data}$ (file-tree, formats, headers, modality cues) without full loading by a Profile-Oriented Data \textbf{Inspector}. An Applicability-Aware Fine-grained Metric \textbf{Selector} then activates feasible metrics and outputs an Evaluation Manifest $\mathcal{M}_{eval}$. Finally, a Knowledge-Augmented Evaluation Specification \textbf{Planner} instantiates each active element $e\in\mathcal{M}_{eval}$ into an executable specification $\pi(e)=\{\pi_{load},\pi_{eval}\}$, where $\pi_{load}$ defines evidence-aware loading (targeting/sampling) and $\pi_{eval}$ defines deterministic scoring grounded by retrieved Type-I domain constraints and informed by Type-II dataset--paper graph signals.
\paragraph{(II) Adaptive and Scalable Tool-Centric Evaluation.}
Given $\pi(e)$, the executor operationalizes evaluation via a structured \textbf{Tool Library} $\mathcal{L}$ and persistent \textbf{Tool Memory} $\mathcal{M}_{tool}$. For each element, it applies Adaptive \textbf{Tool Orchestration} (\emph{retrieve--invoke--synthesize}): retrieve compatible tools, generate invocation code to run them on the loaded evidence, or synthesize and register new tools under cold-start. A \textbf{Review-Driven Verification} layer (runtime + semantic checks) triggers self-correction and, when needed, feeds back to the planner for strategy-level refinement.
Finally, we normalize tool outputs into a unified score format and aggregate them hierarchically (element $\rightarrow$ sub-dimension $\rightarrow$ dimension) to derive multi-dimensional quality scores.

\subsection{Dataset-Aware Evaluation Specification Generation with Knowledge Augmentation}
To accommodate heterogeneous datasets, we develop a domain-knowledge retrieval–augmented method for generating evaluation specifications. Given a target dataset, the method (i) constructs a dataset profile, (ii) selects applicable fine-grained evaluation metrics using an Applicability-Aware Metric Selector, and (iii) precisely defines \emph{what evidence to load} and \emph{how to score} each metric in a deterministic and scientifically grounded manner via a Knowledge-Augmented Evaluation Specification Planner.

\subsubsection{Profile-Oriented Data Inspector}
The Inspector functions as a front-end module that constructs a lightweight \textbf{Data Profile} $\mathcal{P}_{data}$ without fully loading the dataset. It scans the target path to extract the file-tree structure, file formats (e.g., \texttt{.csv}, \texttt{.pdb}, \texttt{.nc}), candidate primary structured files, and available metadata headers. Based on these observations, it infers the dataset modality (e.g., tabular, image, sequence, or multimodal) and summarizes salient characteristics, including storage organization, schema cues, and potential variable or attribute signatures. This metadata-first design improves scalability: by prioritizing header-level and structural probing, the Inspector remains practical for large scientific corpora where full loading is prohibitively expensive.

\subsubsection{Applicability-Aware Fine-grained Metric Selector}
Conditioned on $\mathcal{P}_{data}$, the Selector maps the observed dataset profile to a hierarchical taxonomy of evaluation metrics and activates only those fine-grained metrics that are both meaningful and feasible for the dataset. Leveraging retrieved Type-I knowledge, it determines the \emph{applicability} of each metric and prunes irrelevant branches before triggering any expensive data loading or tool execution. For example, if the Inspector detects paired image--text artifacts, the Selector activates \texttt{ModalAlignment}; otherwise, that branch is disabled. The module outputs an Evaluation Manifest $\mathcal{M}_{eval}$, i.e., a dataset-conditioned set of active metrics to be evaluated. This early-stage filtering reduces unnecessary execution and ensures that subsequent planning is grounded in relevant evaluation objectives.

\subsubsection{Knowledge-Augmented Evaluation Specification Planner}
The Planner is the core component that converts the activated metric set into executable evaluation specifications. Conditioned on the data profile $\mathcal{P}_{data}$, the Evaluation Manifest $\mathcal{M}_{eval}$, and retrieved knowledge, it generates an element-level plan $\pi(e)=\{\pi_{load},\pi_{eval}\}$ for each active element $e$, where $\pi_{load}$ specifies \emph{what evidence to load} and $\pi_{eval}$ specifies \emph{how to score} the metric deterministically. To support domain adaptation, we first construct a knowledge base and then describe the Planner’s multi-stage specification procedure.

\vspace{0.5\baselineskip} 
\noindent\textbf{Knowledge Base Construction.}
Our Knowledge Base (KB) supports both domain-grounded metric instantiation and task-aware dataset reasoning via two components. \textbf{Type-I} is a semi-structured repository indexed by $(\texttt{Domain}, \texttt{Element})$ that records assessment-critical fields (e.g., standards, unit conventions, failure modes, and thresholds); the Planner retrieves it to specialize generic metrics into domain-specific criteria and to constrain $\pi_{load}$ and $\pi_{eval}$ for scientific validity and implementability. \textbf{Type-II} is a dataset--paper graph mined from scientific corpora and linked to open-science repositories (e.g., Hugging Face), encoding dataset--task--model--paper relations; given a dataset $D$, \ModelEvalName~uses it to retrieve similar datasets and summarize downstream use cases, enabling \emph{AI compatibility} and \emph{scientific adaptability} analysis that informs task recommendations and diagnosis of low-scoring criteria.

\vspace{0.5\baselineskip} 
\noindent\textbf{Planning Procedure.}
For each active element $e\in\mathcal{M}_{eval}$, the Planner generates $\pi(e)$ in three stages as follows:

\textbf{Stage 1}: Evidence-Aware Loading Strategy.
Scientific datasets are often too large for full loading and may require specialized parsers. The Planner therefore constructs $\pi_{load}$ by balancing adequacy (sufficient evidence for assessment) and efficiency (minimal I/O). The strategy specifies (i) \emph{targeting}---which artifacts to load (e.g., metadata only, selected variables/columns, or structured components such as coordinates) and (ii) \emph{sampling}---how to select representative subsets (e.g., random sampling for distributional checks, time-slice sampling for temporal consistency, or stratified sampling for rare-event coverage). The resulting specification is designed to be consumed by the execution for code generation and reuse.

\textbf{Stage 2}: Deterministic Evaluation Strategy.
Given the evidence exposed by $\pi_{load}$, the Planner instantiates $\pi_{eval}$ as a deterministic quantification recipe that produces a score in $[0,1]$ together with interpretable quantities. Generic metric descriptions are grounded into dataset- and domain-specific criteria via Type-I retrieval. For example, in climate datasets, the generic check of ``unit consistency'' is instantiated as CMIP-style validation (e.g., verifying that temperature variables use Kelvin). This grounding yields evaluation strategies that are both computable and scientifically defensible.

\textbf{Stage 3}: Strategy-Level Refinement.
\ModelEvalName~maintains a semantic feedback loop between planning and execution. Beyond syntactic or runtime errors handled by tools, \emph{semantic failures} (e.g., null outputs or scores outside $[0,1]$) trigger \emph{strategic reflection}. The Planner attributes the failure to either (a) $\pi_{load}$ (e.g., incorrect file patterns, incompatible parsing, or insufficient sampling) or (b) $\pi_{eval}$ (e.g., invalid assumptions or missing domain constraints), and revises $\pi(e)$ before requesting additional tool actions.

\subsection{Adaptive and Scalable Tool-Centric Evaluation}
\label{sec:tool-executor}
Given $\pi(e)$, the execution subsystem operationalizes evaluation as a tool-centric pipeline that supports tool retrieval, adaptive invocation, and cold-start synthesis. It integrates a structured Tool Library $\mathcal{L}$, a persistent Tool Memory $\mathcal{M}_{tool}$, and a generation--verification loop. Rather than treating execution as a fixed runtime, \ModelEvalName~casts it as a continual-learning process that expands executable coverage across diverse scientific datasets.

\subsubsection{Structured Tool Library Construction}
The Tool Library stores reusable evaluation tools as atomic functional units with rich metadata: (i) identity and scope (a unique tool ID and associated evaluation elements), (ii) operational constraints (accepted input types and domain applicability), (iii) semantic description (method summary and referenced standards), and (iv) encapsulated implementation (self-contained Python code with dependencies imported within the function). This design improves tool management and safe reuse while decoupling planning semantics from implementation details.

\subsubsection{Adaptive and Extensible Tool Orchestration}
To execute $\pi_{eval}$, the Executor uses a three-stage procedure:
(i) \emph{Retrieve}: apply hard filtering to keep tools compatible with the target element and the loaded artifacts, then perform LLM-based semantic selection over tool descriptions under the constraints of $\pi_{eval}$;
(ii) \emph{Invoke}: generate invocation code that binds the loaded data to the selected tool interface, with bounded retries to resolve parameterization errors;
(iii) \emph{Synthesize}: if no suitable tool is found (cold start), synthesize a new tool from $\pi_{eval}$ and the loaded-data profile, verify it, and register it in $\mathcal{L}$.
This orchestration favors reuse for efficiency while enabling synthesis to cover long-tail elements.

\subsubsection{Tool Memory-Guided Execution}
We maintain a persistent Tool Memory $\mathcal{M}_{tool}$ that records execution trajectories and is retrieved as contextual constraints during retrieval, invocation, and synthesis. It contains three streams: (i) \emph{data-loading logs} (successful parsing and sampling patterns), (ii) \emph{success trajectories} (correct invocation exemplars for specific tools and data types), and (iii) \emph{failure trajectories} (runtime errors and semantic mismatches). Success trajectories provide few-shot guidance for parameterization, whereas failure trajectories act as negative constraints that reduce errors.

\subsubsection{Review-Driven Verification and Self-Correction}
The pipeline includes a two-layer review mechanism. \emph{Runtime review} diagnoses code-level failures and guides self-correction (e.g., separating tool-internal faults from invocation errors). \emph{Semantic review} checks whether outputs satisfy $\pi_{eval}$, including score-range constraints ($[0,1]$), required output schema, and domain-consistent values. Semantic violations trigger tool regeneration or strategy-level reflection to the Planner, forming a closed-loop correction channel.


Finally, we synthesizes element-level outputs into a cohesive scientific report. It normalizes tool outputs into a unified score representation and aggregates them hierarchically (element $\rightarrow$ sub-dimension $\rightarrow$ dimension) to form a multi-dimensional score vector. 
Beyond numeric scoring, \ModelEvalName~leverages Type-II Dataset--Paper Pairs to infer latent potential: it traverses the relational graph to identify downstream AI tasks and experimental settings in which structurally similar datasets have been successfully used.
Accordingly, the resulting \textbf{AI-Ready Report} not only diagnoses deficiencies and recommends remediation steps, but also suggests plausible AI-for-Science applications by explicitly linking low scores to constraints documented in prior successful studies (e.g., missing metadata fields required by a benchmark pipeline).

%% file: 6_Experiments.tex
\section{Experiments}
\label{sec:experiments}

We evaluated the performance of our \ModelName on a diverse collection of heterogeneous scientific datasets. We first introduce the dataset preparation and implementation details. Subsequently, we present the main evaluation results, identifying key patterns in scientific data readiness, and provide a quantitative analysis of the system's effectiveness and tool generation capabilities.

\subsection{Experimental Setup}

\subsubsection{Datasets Construction}
To evaluate the generalizability of our framework across heterogeneous scientific contexts, we collected approximately 80 datasets from authoritative open-science repositories, including ScienceDB, Zenodo, Figshare, etc. These datasets span six distinct domains:\textit{Astronomy}, \textit{Biomedicine}, \textit{Earth Science},  \textit{Materials Chemistry}, \textit{Physics \& Engineering},   and \textit{Socio-economics}. During collection, we prioritized diversity across both domain-specific governance standards and data modalities, ensuring comprehensive coverage of tabular and textual files, tensors and graphs, images, and several domain-specific modalities. This heterogeneity serves as a rigorous testbed for the system's adaptability in addressing diverse real-world challenges.

\subsubsection{Implementation Details}
Our multi-agent framework uses gemini-3-pro\footnote{https://deepmind.google/models/gemini/pro} as the backbone LLM. The execution environment for the generated evaluation tools is dynamic: standard evaluation scripts run in a local Python environment, while tools that require complex or conflicting dependencies are isolated in sandboxed Docker containers to ensure stability and security.

\subsubsection{Metrics}
\label{sec:metrics}
In addition to the dataset-level evaluation based on \ModelPrinciplesName~in Section~\ref{sec:tqa2}, we further design a set of metrics to systematically assess the reliability, scalability, and efficiency of \ModelEvalName. Specifically, we consider:

\begin{itemize}[leftmargin=*]
    \item \textbf{System Correctness (SC).}
    Percentage of datasets that successfully yield a complete, multi-dimensional evaluation report without any metric failures within the allowed number of debugging iterations, reflecting overall robustness.


    \item \textbf{Tool Creation Success Rate (TCSR).}
    Fraction of tool-synthesis attempts that yield executable tools whose outputs pass semantic verification, measuring the reliability of agentic tool generation.

    \item \textbf{Tool Creation Efficiency (TCE).}
    Average number of debugging iterations for successfully synthesized tools to pass verification, where lower values indicate more efficient tool generation.
\end{itemize}

\subsection{Main Results}

\input{tables/leaderboards_6domains}

Table~\ref{tab:leaderboard_6domains} presents the quantitative evaluation results for the top-performing datasets across six scientific domains. 
We observe that scores for Governance Trustworthiness ($T$) and Data Quality ($Q$) generally surpass those for AI Compatibility ($A^c$) and Scientific Adaptability ($A^s$).
This trend suggests that while open-science datasets often meet traditional stewardship standards regarding formatting and compliance, they frequently lack the semantic richness or structural alignment required for advanced AI modeling and mechanism discovery.
Notably, the symbol ``-'' indicates metrics automatically pruned by the \textit{Applicability-Aware Selector} as irrelevant to specific modalities or tasks (e.g., class balance for unsupervised learning or spatial coverage for non-geometric data), validating the framework's capacity to conduct context-dependent evaluation rather than applying rigid, static rules.

To validate the semantic correctness of these quantitative assessments, we cross-referenced the generated diagnostic reports with established findings in domain literature. 
For example, the framework's high  $A^s$-score for\textit{Ionospheric\_Obs}\cite{75cb912ede2b4edd99aa78e3208c3c78} aligns with its status as a benchmark for analyzing rare ``superstorm'' events in recent space physics studies~\cite{advancingsustainabledevelopment2025zhao}. 
Crucially, the agent's specific critique of ``tensor architecture mismatches'' accurately reflects the practical engineering hurdles documented in data fusion methodologies, where converting heterogeneous meshgrids to AI-ready tensors remains a manual prerequisite.
Similarly, for \textit{BigSolDB v2.0}~\cite{accuratelypredictingsolubility2025alibrahimb}, our framework accurately decoupled its exceptional Scientific Adaptability ($S_{\text{scale}}=100.0$) from its latent data noise; the agent's specific diagnosis of ``inter-laboratory consistency conflicts'' precisely captures the intrinsic variance introduced by aggregating experimental results from nearly 1,600 diverse literature sources.
Comprehensive evidence mapping specific agent diagnoses to literary corroborations is provided in Table~\ref{tab:evidence_dataset_part1}--Table~\ref{tab:evidence_dataset_part3} in the Appendix. 

We further conducted a blinded human expert validation to assess evaluation correctness. Specifically, we sampled 15 datasets from three domains (Biomedicine, Socio-economics, and Earth Science; 5 datasets per domain) and invited 9 AI-for-Science experts, with 3 experts per domain, to evaluate sub-dimension-level outputs. For each evaluated fragment, experts rated (i) \textit{Score Accuracy}, i.e., whether the assigned score faithfully reflects the dataset on the corresponding sub-dimension, and (ii) \textit{Analysis Quality}, i.e., whether the accompanying diagnostic analysis is reasonable and well justified, using a 5-point Likert scale. We compared \ModelName{} with two ablation variants, w/o Knowledge and w/o Memory, under a blinded setting. As shown in Table~\ref{tab:human_validation}, \ModelName{} achieved higher Score Accuracy and Analysis Quality than both w/o Knowledge and w/o Memory variants, with good inter-rater agreement (ICC=0.742), supporting the reliability of the expert judgments.
\begin{table}[t]
\centering
\caption{Blinded human expert validation on 15 datasets.}
\label{tab:human_validation}
\small
\begin{tabular*}{\columnwidth}{@{\extracolsep{\fill}}lcc@{}}
\toprule
System & Score Accuracy & Analysis Quality \\
\midrule
\ModelName{} & \textbf{4.15} & \textbf{4.11} \\
w/o Knowledge & 3.52 & 3.26 \\
w/o Memory & 3.99 & 3.90 \\
\bottomrule
\end{tabular*}
\end{table}

\subsection{Performance Analysis of \ModelEvalName}
\label{sec:system_evaluation}

To further demonstrate the effectiveness of the proposed \ModelEvalName, we conducted quantitative evaluations based on the evaluation metrics introduced in Section~\ref{sec:metrics}. Additionally, we performed an ablation study to validate the necessity of its architectural components. As illustrated in Figure~\ref{fig:ablation}, the full \ModelEvalName demonstrates robust overall performance, achieving a framework evaluation success rate of 89.0\% and a tool creation success rate of 97.4\%, while maintaining a high generation efficiency with an average of only 1.19 iterations per tool. 

Additional robustness analyses, including backbone LLM comparison, baseline comparison with existing tools, and scalability/cost analysis, are reported in Appendix~\ref{app:additional_analyses}.

To verify the necessity of each architectural component, we conducted an ablation study by systematically degrading key modules. We compared \ModelName~against three variants:
\begin{itemize}[leftmargin=*]
    \item \textbf{w/o Knowledge Planning}: We removed the knowledge retrieval module from the \textit{Planner}, forcing it to generate evaluation strategies based solely on the LLM's parametric knowledge. Crucially, the \textit{Reviewer} retained access to domain knowledge (Type-I) to verify the results. 
    \item \textbf{w/o Tool Memory}: We disabled the retrieval of existing tools and historical "success trajectories," forcing the system to synthesize every tool \textit{de novo} without few-shot guidance.
    \item \textbf{w/o Self-Correction}: We disabled the \textit{Review-Driven Verification and Self-Correction}, restricting the tool generation to a single-pass attempt ($K=1$) without debug iterations.
\end{itemize}

The ablation results highlight the distinct contributions of each module. 
First, the removal of the \textit{Iterative Review} mechanism causes the most significant performance degradation, precipitating a sharp drop in framework evaluation success to 33.0\%. This indicates that the self-correction loop is critical for resolving runtime errors and ensuring the executability of synthesized tools.
Second, the \textit{w/o Knowledge} variant exhibits a substantial decline in evaluation success (51.7\%) and records the lowest tool creation efficiency (1.48 iterations). This suggests that without retrieved domain-knowledge constraints, the system struggles to generate valid evaluation logic efficiently, leading to higher trial-and-error costs.
Finally, the \textit{w/o Memory} setting results in a moderate performance decrease (Evaluation Success: 82.6\%, Efficiency: 1.30). While the system remains functional, this confirms that the tool memory facilitates stable generation and reduces computational overhead by retrieving proven execution patterns.

\subsection{Case Study}
\label{sec:case_study}

\begin{figure}
    \centering
    \includegraphics[width=\linewidth]{figs/ablation_comparison_combined.pdf}
    \caption{The performance of Sci-TQA²-Eval and its variants.}
    \label{fig:ablation}
\end{figure}

To illustrate the automated evaluation workflow of \ModelEvalName, we present a representative case study on the DiTing~\cite{ZHAO202384}, a large-scale dataset for seismic analysis. As depicted in Figure~\ref{fig:case_study_diting}, the analysis focuses on the \textit{Target Distribution Health} dimension, which determines the statistical suitability of ground truth labels for model convergence. The process begins with the \textbf{Planner} retrieving domain-specific knowledge, correctly identifying that seismic magnitudes inherently follow the exponential Gutenberg-Richter law~\cite{gutenberg1956earthquake}. Subsequently, the \textbf{Executor} invokes the statistical assessment tool to quantify distributional properties, calculating skewness penalties for regression targets and entropy for classification labels. The resulting score of 55.2/100 accurately captures the characteristics of the evaluated data subset: while the distribution is physically valid, the sampled partition exhibits a heavy-tailed skew and class imbalance that may destabilize standard loss functions. Finally, based on these diagnostics, the system synthesises a report that maps the dataset to appropriate downstream tasks (e.g., Phase Picking) and recommends robust model architectures~\cite{mousavi2020earthquake}.

%% file: tables/leaderboards_6domains.tex
\begin{table*}[t]
\footnotesize
\setlength{\tabcolsep}{4pt}
\vspace{3mm}
\caption{Quantitative Evaluation Results — All Domains (Part I). Background intensity correlates with score magnitude.
\textbf{$T$}: Governance Trustworthiness ($T_1$: FAIRness, $T_2$: Provenance, $T_3$: Ethics);
\textbf{$Q$}: Data Quality ($Q_1$: Complete., $Q_2$: Acc., $Q_3$: Unique., $Q_4$: Consist.);
\textbf{$A^c$}: AI Compatibility ($A^c_1$: Adapt., $A^c_2$: Scale, $A^c_3$: Bal., $A^c_4$: Feat.Imp.);
\textbf{$A^s$}: Scientific Adaptability ($A^s_1$: Task Gen., $A^s_2$: Scarcity, $A^s_3$: Causal, $A^s_4$: Cond. Cov.).
\textit{Modality}: \faGlobe~Tensor, \faProjectDiagram~Graph, \faTable~Table, \faImage~Image, \faFont~Text.}
\label{tab:leaderboard_6domains}

\centering
\begin{tabular*}{\textwidth}{@{\extracolsep{\fill}} l c c | ccc | cccc | cccc | cccc}
\toprule
\multirow{2}{*}{\textbf{Dataset}} &
\multirow{2}{*}{\textbf{Mod.}} &
\multirow{2}{*}{\textbf{Total}} &
\multicolumn{3}{c|}{\textbf{Gov. Trust. ($T$)}} &
\multicolumn{4}{c|}{\textbf{Data Qual. ($Q$)}} &
\multicolumn{4}{c|}{\textbf{AI Comp. ($A^c$)}} &
\multicolumn{4}{c}{\textbf{Sci. Adapt. ($A^s$)}} \\
\cmidrule(lr){4-6} \cmidrule(lr){7-10} \cmidrule(lr){11-14} \cmidrule(lr){15-18}
 &  &  & $T_1$ & $T_2$ & $T_3$ & $Q_1$ & $Q_2$ & $Q_3$ & $Q_4$ & $A^c_1$ & $A^c_2$ & $A^c_3$ & $A^c_4$ & $A^s_1$ & $A^s_2$ & $A^s_3$ & $A^s_4$ \\
\midrule

\rowcolor{blue!8} \multicolumn{18}{c}{\textit{\textbf{Astronomy}}} \\
\midrule
Ring\_Moat\_Dome\_Str. & \faImage & 88.6 & \gGood 76.0 & \gHigh 100.0 & \gHigh 100.0 & \gHigh 100.0 & \gHigh 100.0 & \gHigh 100.0 & \gHigh 99.8 & \gMid 67.5 & \gHigh 100.0 & \gMid 73.4 & \gLow 45.7 & \gHigh 93.8 & \gHigh 93.3 & \gGood 89.0 & \gGood 87.0 \\
Ionospheric\_Obs. & \faGlobe & 88.1 & \gGood 76.0 & \gHigh 100.0 & \gHigh 100.0 & \gGood 86.8 & \gGood 82.4 & \gMid 74.3 & \gHigh 100.0 & \gGood 82.8 & \gGood 78.8 & - & \gBad - & \gHigh 100.0 & \gHigh 100.0 & \gGood 87.6 & \gGood 87.4 \\
Martian\_Aeol.\_Land. & \faImage & 86.2 & \gGood 76.0 & \gHigh 100.0 & \gHigh 100.0 & \gGood 75.0 & \gHigh 100.0 & \gHigh 100.0 & \gHigh 100.0 & \gHigh 93.7 & \gHigh 100.0 & \gMid 61.3 & - & \gGood 87.5 & \gGood 86.0 & \gLow 45.3 & \gGood 78.0 \\
SOLAR\_WIND\_MODEL & \faTable & 83.7 & \gGood 80.0 & \gHigh 100.0 & \gHigh 100.0 & \gHigh 98.0 & \gHigh 90.9 & \gHigh 99.9 & \gHigh 100.0 & \gGood 87.5 & \gHigh 99.3 & \gGood 82.0 & \gBad 15.4 & \gHigh 100.0 & \gHigh 97.0 & \gHigh 93.7 & \gBad 24.2 \\
Solar\_Act.\_AI\_Fore. & \faImage,\faTable & 81.8 & \gGood 76.0 & \gHigh 100.0 & \gHigh 100.0 & \gHigh 93.5 & \gGood 88.5 & \gBad 0.0 & \gHigh 100.0 & \gMid 73.3 & \gHigh 100.0 & \gGood 83.4 & \gBad - & \gHigh 100.0 & \gHigh 94.5 & \gMid 61.3 & \gLow 46.7 \\

\midrule
\rowcolor{green!8} \multicolumn{18}{c}{\textit{\textbf{Biomedicine}}} \\
\midrule
Flysta3D & \faImage & 91.4 & \gGood 85.7 & \gHigh 100.0 & \gGood 79.2 & \gHigh 100.0 & \gHigh 100.0 & \gHigh 100.0 & \gHigh 100.0 & \gHigh 99.6 & \gMid 71.2 & - & \gHigh 100.0 & \gHigh 98.1 & \gHigh 100.0 & \gMid 53.3 & - \\
The-Gonzo-Dataset & \faGlobe,\faTable & 86.8 & \gHigh 100.0 & \gMid 55.6 & \gGood 83.3 & \gHigh 100.0 & \gLow 48.8 & - & \gHigh 100.0 & \gHigh 100.0 & \gGood 78.3 & - & - & \gHigh 100.0 & \gHigh 100.0 & \gGood 82.7 & \gHigh 100.0 \\
Synth.-Skeletal-Mot. & \faImage,\faTable & 81.4 & \gHigh 100.0 & \gLow 44.4 & \gGood 83.3 & \gHigh 100.0 & \gMid 50.0 & \gHigh 96.8 & \gHigh 100.0 & \gGood 87.7 & \gMid 68.0 & \gGood 84.5 & \gHigh 91.0 & \gHigh 92.2 & \gMid 60.0 & \gGood 86.3 & \gGood 87.1 \\
MedMNISTv2 & \faGlobe & 80.2 & \gHigh 100.0 & \gMid 55.6 & \gGood 83.3 & \gHigh 100.0 & \gHigh 100.0 & \gHigh 92.1 & \gHigh 100.0 & \gMid 70.0 & \gGood 75.6 & \gGood 80.3 & - & \gGood 85.0 & \gMid 74.7 & \gBad 6.0 & \gGood 85.2 \\
Genomic-benchmarks & \faFont & 79.9 & \gMid 60.0 & \gMid 66.7 & \gMid 66.7 & \gHigh 90.6 & \gGood 78.1 & \gMid 70.0 & \gHigh 100.0 & \gHigh 94.0 & \gGood 79.9 & \gHigh 98.6 & - & \gMid 62.5 & \gGood 80.0 & \gBad 0.0 & \gHigh 96.1 \\

\midrule
\rowcolor{cyan!8} \multicolumn{18}{c}{\textit{\textbf{Earth Science}}} \\
\midrule
Fengyun-3 & \faImage & 90.2 & \gGood 76.0 & \gHigh 100.0 & \gHigh 100.0 & \gHigh 93.5 & \gGood 88.0 & \gHigh 100.0 & \gHigh 100.0 & \gMid 73.5 & \gHigh 91.5 & - & - & \gHigh 99.8 & \gHigh 100.0 & \gMid 63.3 & \gHigh 100.0 \\
Space\_Hurr.\_Recog. & \faImage & 86.0 & \gGood 76.0 & \gHigh 100.0 & \gHigh 100.0 & \gHigh 100.0 & \gGood 88.3 & \gLow 45.8 & \gMid 62.9 & \gHigh 90.0 & \gHigh 93.1 & \gHigh 100.0 & - & \gGood 81.5 & \gHigh 93.3 & \gHigh 90.0 & \gHigh 90.4 \\
Turb.\_Heat\_Flux\_2023 & \faGlobe & 85.7 & \gGood 76.0 & \gHigh 100.0 & \gHigh 100.0 & \gGood 78.1 & \gGood 85.2 & \gHigh 100.0 & \gHigh 100.0 & \gHigh 94.0 & \gGood 80.0 & - & \gHigh 100.0 & \gMid 50.4 & \gHigh 95.0 & \gMid 60.0 & \gBad 10.0 \\
LoveDA & \faImage & 85.5 & \gHigh 100.0 & \gMid 55.6 & \gGood 83.3 & \gHigh 100.0 & \gGood 89.3 & \gHigh 100.0 & \gHigh 100.0 & \gGood 87.5 & \gGood 89.8 & \gHigh 94.0 & - & \gGood 77.5 & \gMid 71.7 & \gMid 55.0 & \gHigh 93.9 \\
EM\_Field\_Structure & \faGlobe & 85.2 & \gGood 76.0 & \gHigh 100.0 & \gHigh 100.0 & \gHigh 100.0 & - & - & \gHigh 100.0 & \gHigh 99.9 & \gMid 51.0 & - & - & \gMid 70.8 & \gGood 80.0 & \gGood 84.0 & \gMid 58.5 \\

\midrule
\rowcolor{orange!8} \multicolumn{18}{c}{\textit{\textbf{Materials \& Chemistry}}} \\
\midrule
FreeSolv & \faTable,\faProjectDiagram & 88.2 & \gGood 85.7 & \gHigh 100.0 & \gGood79.2 & \gHigh 100.0 & \gHigh 100.0 & \gHigh 100.0 & \gHigh 99.8 & \gHigh 100.0 & \gMid 56.2 & \gGood 76.2 & \gHigh 96.8 & \gGood 87.5 & \gHigh 97.5 & \gHigh 100.0 & \gHigh 90.3 \\
BigSolDBv2.0 & \faTable & 88.0 & \gHigh 100.0 & \gMid 55.6 & \gGood 83.3 & \gHigh 99.3 & \gHigh 91.3 & \gHigh 100.0 & \gHigh 100.0 & \gGood 89.2 & \gHigh 100.0 & \gGood 80.1 & \gGood 77.6 & \gMid 68.8 & \gHigh 100.0 & \gHigh 95.0 & \gGood 87.5 \\
xxMD & \faProjectDiagram & 86.3 & \gGood 76.0 & \gHigh 100.0 & \gHigh 100.0 & \gHigh 100.0 & \gHigh 99.3 & \gHigh 100.0 & \gHigh 100.0 & \gHigh 90.6 & \gGood 84.0 & \gMid 71.8 & \gLow 47.6 & \gHigh 100.0 & \gHigh 100.0 & \gLow 48.2 & \gGood 85.4 \\
Raw\_Hyperspec.\_Refl. & \faTable & 85.0 & \gHigh 96.0 & \gHigh 100.0 & \gHigh 95.8 & \gHigh 99.8 & \gHigh 99.4 & \gHigh 100.0 & \gHigh 100.0 & \gGood 81.2 & \gBad 16.0 & \gGood 87.2 & \gMid 73.7 & \gMid 71.2 & \gHigh 100.0 & \gMid 66.7 & \gGood 81.9 \\
1018\_Compounds & \faTable & 85.0 & \gHigh 96.0 & \gHigh 100.0 & \gHigh 95.8 & \gHigh 100.0 & \gMid 71.0 & \gHigh 100.0 & \gHigh 100.0 & \gHigh 99.9 & \gMid 68.1 & \gHigh 99.4 & - & \gMid 62.5 & \gHigh 90.0 & \gLow 40.0 & \gMid 72.0 \\

\midrule
\rowcolor{red!8} \multicolumn{18}{c}{\textit{\textbf{Physics
\& Engineering}}} \\
\midrule
Multiharm.\_Feedback & \faTable & 93.2 & \gHigh 100.0 & \gHigh 100.0 & \gHigh 100.0 & \gHigh 100.0 & \gHigh 100.0 & \gHigh 100.0 & \gHigh 100.0 & \gHigh 100.0 & \gGood 83.9 & - & \gGood 82.4 & \gMid 66.7 & \gGood 85.0 & \gGood 85.3 & \gHigh 100.0 \\
CdS\_Nanowire\_SHG & \faTable & 88.9 & \gHigh 92.0 & \gHigh 100.0 & \gHigh 100.0 & \gHigh 100.0 & \gMid 55.6 & \gHigh 100.0 & \gHigh 100.0 & \gHigh 96.0 & \gLow 49.4 & - & \gGood 86.6 & \gGood 87.5 & \gHigh 90.0 & \gHigh 90.0 & \gHigh 100.0 \\
Dusty\_Plasma\_Ratchet & \faTable & 86.8 & \gHigh 96.0 & \gHigh 100.0 & \gHigh 100.0 & \gHigh 100.0 & - & \gHigh 100.0 & \gHigh 92.9 & \gGood 86.2 & \gMid 70.3 & \gMid 55.5 & \gHigh 94.5 & \gHigh 93.8 & \gLow 25.0 & \gGood 82.9 & \gGood 76.7 \\
Tokamak\_Dust\_Dynamics & \faTable & 85.1 & \gGood 76.0 & \gHigh 100.0 & \gHigh 100.0 & \gHigh 100.0 & \gHigh 100.0 & \gHigh 100.0 & \gHigh 100.0 & \gMid 65.1 & \gMid 55.0 & - & \gHigh 100.0 & \gMid 70.0 & \gHigh 100.0 & \gGood 83.3 & \gLow 47.0 \\
Coherent\_Light\_Int. & \faTable & 85.0 & \gGood 76.0 & \gHigh 100.0 & \gHigh 100.0 & \gHigh 100.0 & \gGood 80.0 & \gHigh 100.0 & \gHigh 100.0 & \gHigh 100.0 & \gMid 65.0 & - & - & \gGood 81.2 & \gMid 70.0 & \gGood 76.5 & \gMid 55.0 \\

\midrule
\rowcolor{yellow!8} \multicolumn{18}{c}{\textit{\textbf{Socio-economics}}} \\
\midrule
Dig.\_Econ.\_Green\_Dev. & \faTable & 95.3 & \gHigh 100.0 & \gHigh 100.0 & \gHigh 100.0 & \gHigh 100.0 & - & \gHigh 100.0 & \gHigh 100.0 & \gGood 82.5 & \gHigh 93.1 & \gBad - & \gGood 85.3 & \gHigh 100.0 & \gHigh 100.0 & \gHigh 96.7 & \gGood 81.0 \\
Env.\_Reg.\_Green\_Inn. & \faTable & 91.0 & \gHigh 100.0 & \gHigh 100.0 & \gHigh 100.0 & \gHigh 100.0 & - & \gHigh 100.0 & \gHigh 100.0 & \gHigh 100.0 & \gGood 87.5 & - & \gMid 66.0 & \gHigh 100.0 & \gMid 56.8 & \gHigh 95.3 & \gMid 57.6 \\
China\_Cross\_Bord.\_Ecom. & \faTable & 89.2 & \gHigh 100.0 & \gHigh 100.0 & \gHigh 100.0 & \gMid 74.1 & \gHigh 100.0 & \gHigh 100.0 & \gHigh 100.0 & \gHigh 99.8 & \gHigh 96.4 & \gLow 49.4 & \gMid 72.6 & \gHigh 100.0 & \gHigh 99.7 & \gHigh 100.0 & \gGood 85.0 \\
China\_282\_Green\_Econ. & \faTable & 88.6 & \gGood 76.0 & \gHigh 100.0 & \gHigh 100.0 & \gHigh 100.0 & - & \gHigh 100.0 & \gHigh 100.0 & \gGood 75.0 & \gGood 88.6 & \gMid 71.3 & \gGood 88.1 & \gHigh 100.0 & \gGood 77.6 & \gGood 75.0 & \gGood 87.3 \\
Mgmt.\_Tone\_Stock\_Pr. & \faTable & 87.5 & \gGood 80.0 & \gGood 88.9 & \gHigh 100.0 & \gHigh 100.0 & \gMid 50.0 & \gHigh 100.0 & \gHigh 100.0 & \gGood 89.1 & \gMid 73.4 & - & \gGood 87.4 & \gHigh 100.0 & \gHigh 95.0 & \gGood 75.8 & \gGood 86.7 \\
W.\_China\_Farmer\_Surv. & \faTable & 86.7 & \gHigh 100.0 & \gHigh 100.0 & \gHigh 100.0 & \gHigh 98.0 & - & \gHigh 100.0 & \gHigh 99.9 & \gHigh 90.2 & \gLow 45.6 & \gHigh 100.0 & \gHigh 98.6 & \gGood 76.7 & \gHigh 90.0 & \gLow 37.0 & - \\

\bottomrule
\end{tabular*}
\end{table*}

%% file: 7_Conclusion.tex
\section{Limitations and Ethical Considerations}
Our method was evaluated using open-source Heterogeneous Scientific Datasets from six scientific domains. While these datasets span various fields, their scope remains somewhat limited. Future work could include datasets from additional domains to further validate and expand the method’s applicability. Regarding ethics, all datasets used in this study are publicly available, and we have strictly followed the usage guidelines for each. These datasets were used solely for validating our proposed method, ensuring that no ethical concerns arise from their use. Another limitation is that the current AI Compatibility assessment is grounded in known model architectures; future work will extend the taxonomy as new AI paradigms emerge to avoid over-favoring any single modeling family. \looseness=-1

\begin{figure}[t]
    \vspace{-2mm}
    \centering
    \begin{tcolorbox}[
        colback=gray!5, 
        colframe=black!70, 
        title=\textbf{Case Study: Evaluation Trace for DiTing (Seismology) },
        fonttitle=\bfseries\small,
        boxrule=0.8pt,
        arc=2mm
    ]
    \scriptsize
    \textbf{\textsc{Phase 1: Knowledge-Augmented Planning}} \\
    \texttt{[INPUT]} Dataset: \textit{DiTing (Subset: part\_27.csv \& part\_27.hdf5, 3-4\% of total)} \\
    \texttt{[Sub-Dimension]} Class Balance\\ 
    \texttt{[RETRIEVAL]} Domain: Seismology $\rightarrow$ Aspect: Target Distribution Health \\ 
    \texttt{[CONSTRAINT]} \textcolor{blue}{Gutenberg-Richter Law}: Earthquake magnitudes follow an exponential decay ($N \propto 10^{-bM}$). \textbf{Expectation}: High skewness in regression targets, dominated by small earthquakes and scarce large events, may hinder the accurate prediction of larger earthquakes if untreated.
    
    \vspace{0.1cm}
    \hrule
    \vspace{0.1cm}
    
    \textbf{\textsc{Phase 2: Adaptive Tool Generation \& Execution (Snapshot)}} \\
    \texttt{[TOOL]} \texttt{assess\_target\_distribution\_health} \\
    \texttt{[LOGIC]} \textit{Calculating statistical suitability for ML training:}
    \begin{lstlisting}[basicstyle=\tiny\ttfamily, breaklines=true, language=Python, keywordstyle=\color{blue}, numbers=none, frame=none]
def assess_distribution(data, target_config):
    if col_type == 'continuous': # e.g., 'evmag'
        # Penalty for exponential distribution (G-R Law detection)
        ......
    elif col_type == 'categorical': # e.g., 'p_motion'
        # Normalized Shannon Entropy for Class Balance
        ......
    return raw_score * (1.0 - nan_ratio)
    \end{lstlisting}
    \texttt{[RESULT]} \texttt{evmag} skewness $\gamma > 2.5$ $\rightarrow$ Score penalized. \\
    \texttt{[AGGREGATED SCORE]} \textbf{55.2 / 100 (Moderate)}
    
    \vspace{0.1cm}
    \hrule
    \vspace{0.1cm}
    
    \textbf{\textsc{Phase 3: AI-Ready Report Generation}} \\
    \texttt{[DIAGNOSIS]} \textbf{Target Distribution Health (55.2/100)}:
    The dataset exhibits an imbalanced magnitude distribution between small and large earthquakes in magnitude targets and class imbalance in polarity labels within this subset. \\
    \texttt{[\textcolor{red}{WARNING}]} Direct training with MSE loss will be dominated by small-magnitude events; Polarity classifiers may suffer from mode collapse. \\
    \texttt{[RECOMMENDED TASKS]} 
    \begin{itemize}[leftmargin=1.5em, noitemsep, topsep=0pt, label=\ding{51}]
        \item \textbf{Seismic Phase Picking}: High suitability due to dense P/S wave labels.
        \item \textbf{Polarity Classification}: Feasible but requires re-balancing class weights.
    \end{itemize}
    \texttt{[MODEL SUGGESTION]} \textbf{EQTransformer} (Hierarchical Attention):
    Recommended to handle the multi-task nature (detection + picking) and mitigate noise in the subset.
    \end{tcolorbox}
    \caption{An example of the evaluation trace and results of \ModelEvalName~ on the DiTing dataset.}
    \label{fig:case_study_diting}
\end{figure}

\section{Conclusion}
In this work, we addressed the growing need for principled and scalable evaluation of AI-readiness in scientific data. We proposed {\ModelName}, an agentic system that systematically assessed the AI-readiness of heterogeneous scientific datasets by jointly considering governance, quality, compatibility, and adaptability factors that are essential for effective learning and scientific reasoning. At the criteria level, we introduced the \ModelPrinciplesName, which decomposed AI-readiness into four complementary dimensions—Governance Trustworthiness, Data Quality, AI Compatibility, and Scientific Adaptability—and further operationalized them through measurable atomic elements. To enable scalable evaluation in practice, we developed \ModelEvalName, a hierarchical multi-agent approach that dynamically constructed and executed dataset-aware evaluation specifications through a directed, cyclic workflow. By integrating lightweight profiling, applicability-aware metric selection, knowledge-augmented planning, and review-driven verification, the proposed system supported reliable and scalable assessment across diverse scientific data modalities. Extensive experiments on heterogeneous datasets spanning multiple scientific domains demonstrated the effectiveness and generality of our approach. 


\begin{acks}
The development of this project was supported by the Scihorizon platform.
\end{acks}

\section{GenAI Disclosure}
In this work, we developed an agentic system based on LLMs to assess the AI-readiness of scientific data. The operation of this system relies on LLMs as core components for reasoning, analysis, and interaction. Additionally, during the preparation of this manuscript, we used generative artificial intelligence–based tools to support language-related tasks, such as grammar checking and improving writing clarity. These tools were not used to generate scientific content, interpret results, or draw conclusions. All scientific ideas, analyses, and interpretations presented in this paper are solely the responsibility of the authors.

%% file: 8_Appendix.tex
\appendix

\setcounter{table}{0}
\setcounter{figure}{0}
\renewcommand{\thetable}{S\arabic{table}}
\renewcommand{\thefigure}{S\arabic{figure}}

\section{Appendix}

\subsection{Details of \ModelPrinciplesName}
\label{app:elements_define}


\paragraph{Governance-related}
\textbf{FAIRness Principles} assess if a dataset is discoverable and usable through machine-actionable metadata, including persistent identifiers, transparent access protocols, interoperable representations, and reuse-oriented documentation.  
\textbf{Provenance \& Licensing} consolidate legal terms and attribution requirements necessary for compliant redistribution and derivative use, thereby reducing ambiguity in downstream model training and publication.  
\textbf{Scientific Ethics} encompasses evidence of responsible data stewardship, including informed consent, privacy protection, and assessment of dual-use risks.

\paragraph{Data-quality}
\textbf{Completeness} measures missingness and broken associations across assets and modalities (e.g., corrupted files or missing cross-references), which directly affect end-to-end training or evaluation.  
\textbf{Consistency} evaluates format stability—such as shape or type mismatches—that can cause parsing failures, batch processing errors, or silent bugs.  
\textbf{Accuracy} reflects alignment of values and labels with scientific and syntactic constraints, including measurement validity and schema compliance.  
\textbf{Uniqueness} identifies duplicate samples and identifier collisions that may inflate performance estimates, induce data leakage, or bias inference.

\paragraph{AI-compatibility}
\textbf{AI Model Adaptability} captures alignment between data topology, task formulation, and model architectures or training strategies, enhancing reproducibility.  
\textbf{Data Scale} indicates whether dataset size is sufficient relative to domain expectations and effective feature dimensionality, directly influencing overfitting risk and generalization.  
\textbf{Class Balance} summarizes the health of target distributions and adequacy of minority-class support for stable learning and reliable metric estimation.  
\textbf{Feature Importance} evaluates feature informativeness and redundancy, guiding decisions on feature selection, dimensionality reduction, or additional data collection.  
\textbf{Data Point Value} quantifies the marginal utility of individual samples, enabling dataset pruning, debugging, and prioritization under labeling budget constraints.

\paragraph{Scientific-adaptability}
\textbf{Task Generalizability} assesses whether the dataset’s schema and semantic coverage support multiple scientific tasks beyond a single benchmark.  
\textbf{Data Scarcity} contextualizes dataset value by capturing barriers to generating comparable data—such as experimental cost, non-repeatable phenomena, or reliance on expert annotation.  
\textbf{Causal Completeness} aggregates signals required to move beyond correlation, including interventional data, process-level resolution, and comprehensive contextual variables to mitigate confounding.  
\textbf{Condition Coverage} measures the breadth of relevant parameter regimes spanned by the data, including phase diversity and boundary or extreme cases, which are critical for extrapolation, robustness, and scientific insight.

\subsection{Additional Experimental Details}

\subsubsection{Metric Definitions}

To quantitatively evaluate the reliability, scalability, and efficiency of \ModelName{}, we define three core metrics. Let $\mathcal{E}_d$ denote the set of atomic evaluation elements for dataset $d$, and $\mathcal{D}$ the set of all evaluated datasets.

\noindent \textbf{1. System Correctness (SC).}  
This metric measures the framework’s overall robustness as the proportion of evaluation elements that yield a valid score within at most five debugging iterations:
\begin{equation}
\text{SC} = \frac{\sum_{d \in \mathcal{D}} \sum_{e \in \mathcal{E}_d} \mathbb{I}(\text{status}(e) = \text{Success})}{\sum_{d \in \mathcal{D}} |\mathcal{E}_d|},
\end{equation}
where $\mathbb{I}(\cdot)$ is the indicator function. A ``Success'' requires both successful execution of the underlying tool (retrieved or synthesized) and passage of semantic verification.

\noindent \textbf{2. Tool Creation Success Rate (TCSR).}  
TCSR evaluates the agent’s ability to synthesize valid, executable tools for previously unseen evaluation tasks. A synthesized tool is deemed valid only if it executes without runtime errors \emph{and} passes semantic verification. Formally:
\begin{equation}
\text{TCSR} = \frac{|\mathcal{T}_{\text{valid}}|}{|\mathcal{T}_{\text{total}}|},
\end{equation}
where $\mathcal{T}_{\text{total}}$ is the set of all synthesis attempts triggered by retrieval failure, and $\mathcal{T}_{\text{valid}}$ is the subset that completes the execution–verification loop within $K$ iterations.

\noindent \textbf{3. Tool Creation Efficiency (TCE).}  
For successfully synthesized tools, TCE measures the average number of debugging iterations required to pass verification:
\begin{equation}
\text{TCE} = \frac{1}{|\mathcal{T}_{\text{success}}|} \sum_{t \in \mathcal{T}_{\text{success}}} \text{Iter}(t).
\end{equation}
Lower TCE indicates higher synthesis precision and reduced computational overhead.

\subsubsection{Additional Robustness Analyses}
\label{app:additional_analyses}

\paragraph{Backbone LLM comparison.}
We evaluated \ModelEvalName{} with alternative backbone LLMs on 18 datasets across all six domains. Table~\ref{tab:backbone_comparison} shows that most strong backbones achieve comparable SC, while the large drop without self-correction confirms that the system design is more critical than the specific backbone.

\begin{table}[h]
\centering
\caption{Backbone LLM comparison on 18 datasets.}
\label{tab:backbone_comparison}
\small
\begin{tabularx}{\columnwidth}{@{}X>{\centering\arraybackslash}X@{}}
\toprule
Backbone & SC (\%) \\
\midrule
Gemini-3-Pro & 89.0 \\
Claude Sonnet 4.6 & 88.9 \\
GLM-5 & 82.5\\
GPT-5.1 & 79.7 \\
DeepSeek-v3.2 & 73.2 \\
Gemini-3-Pro w/o Self-Corr. & 33.0 \\
\bottomrule
\end{tabularx}
\end{table}

\paragraph{Baseline comparison.}
We compared \ModelName{} with two open-source tools, FAIR-Checker and AIDRIN, on the same 15-dataset subset used in the expert validation. FAIR-Checker primarily evaluates metadata encoding compliance, while AIDRIN focuses on tabular data-quality profiling. Existing tools cover only partial dimensions of Sci-TQA²: at least 80\% of our fine-grained sub-dimensions cannot be assessed by either tool, and AIDRIN failed on 8/15 datasets due to format limitations. FAIR-Checker mainly checks RDF-level metadata, while AIDRIN provides generic statistical snapshots; in contrast, \ModelName{} performs context-aware assessment using LLM reasoning and retrieved domain knowledge.


\paragraph{Scalability and cost.}
Across approximately 80 datasets, the average wall-clock time is 22m31s and the average API cost is \$7.22 per evaluation. Cost scales mainly with the number of active evaluation elements rather than raw dataset size, since the Inspector avoids full data loading. The Tool Library also amortizes cold-start synthesis: average evaluation time decreases from 29m49s for the first 20 datasets to 16m35s for the last 20.


\subsubsection{Quantitative Evaluation Results}
\label{app:quant_results}

\input{tables/appendix/all_domains}
We report full evaluation results across six scientific domains: Astronomy, Physics \& Engineering, Biomedicine, Materials \& Chemistry, Earth Science, and Socio-economics.  
For each dataset, we provide (i) an overall score and (ii) a breakdown across four dimensions: Governance Trustworthiness ($T$), Data Quality ($Q$), AI Compatibility ($A^c$), and Scientific Adaptability ($A^s$).  
We retain sub-dimension scores rather than reporting only aggregates, as different downstream applications impose distinct minimum requirements (e.g., licensing compliance vs.\ causal inference vs.\ generalization robustness).

Tables~\ref{tab:leaderboard_appendix_part1} and~\ref{tab:leaderboard_appendix_part2} present all results. Scores are normalized to $[0,100]$ at the sub-dimension level and aggregated upward.  
Cells marked ``--'' indicate that a sub-dimension is \emph{not applicable} (e.g., class balance for unlabeled datasets, or certain AI-compatibility checks for incompatible modalities). We do not impute these entries to avoid conflating inapplicability with poor quality.

\subsubsection{Evidence Collection and Traceability}
\input{tables/appendix/evidence}

To ensure transparency and auditability, we provide evidence tables linking representative scores to justifications in source publications. Due to space constraints, evidence is distributed across Tables~\ref{tab:evidence_dataset_part1}--\ref{tab:evidence_dataset_part3}.  
Each entry includes: (\textbf{Dataset}, \textbf{Dimension}, \textbf{Score}, \textbf{Article}, \textbf{Evidence}).  
The \textbf{Evidence} field contains minimal quoted excerpts used by \ModelName{} to justify the score, enabling manual verification and future re-annotation as scoring rubrics evolve.  
Note that these tables are not exhaustive execution logs; they serve as human-readable traces for selected (dataset, metric) pairs that are either representative or potentially contentious.

%% file: tables/appendix/all_domains.tex
\begin{table*}[t]
\footnotesize
\setlength{\tabcolsep}{4pt}
\caption{Quantitative Evaluation Results — All Domains (Part I). Background intensity correlates with score magnitude.
\textbf{$T$}: Governance Trustworthiness ($T_1$: FAIRness, $T_2$: Provenance, $T_3$: Ethics);
\textbf{$Q$}: Data Quality ($Q_1$: Complete., $Q_2$: Acc., $Q_3$: Unique., $Q_4$: Consist.);
\textbf{$A^c$}: AI Compatibility ($A^c_1$: Adapt., $A^c_2$: Scale, $A^c_3$: Bal., $A^c_4$: Feat.Imp.);
\textbf{$A^s$}: Scientific Adaptability ($A^s_1$: Task Gen., $A^s_2$: Scarcity, $A^s_3$: Causal, $A^s_4$: Cond. Cov.).
\textit{Modality}: \faGlobe~Tensor, \faProjectDiagram~Graph, \faTable~Table, \faImage~Image, \faFont~Text.}
\label{tab:leaderboard_appendix_part1}

\centering
\begin{tabular*}{\textwidth}{@{\extracolsep{\fill}} l c c | ccc | cccc | cccc | cccc}
\toprule
\multirow{2}{*}{\textbf{Dataset}} &
\multirow{2}{*}{\textbf{Mod.}} &
\multirow{2}{*}{\textbf{Total}} &
\multicolumn{3}{c|}{\textbf{Gov. Trust. ($T$)}} &
\multicolumn{4}{c|}{\textbf{Data Qual. ($Q$)}} &
\multicolumn{4}{c|}{\textbf{AI Comp. ($A^c$)}} &
\multicolumn{4}{c}{\textbf{Sci. Adapt. ($A^s$)}} \\
\cmidrule(lr){4-6} \cmidrule(lr){7-10} \cmidrule(lr){11-14} \cmidrule(lr){15-18}
 &  &  & $T_1$ & $T_2$ & $T_3$ & $Q_1$ & $Q_2$ & $Q_3$ & $Q_4$ & $A^c_1$ & $A^c_2$ & $A^c_3$ & $A^c_4$ & $A^s_1$ & $A^s_2$ & $A^s_3$ & $A^s_4$ \\
\midrule

\rowcolor{blue!8} \multicolumn{18}{c}{\textit{\textbf{Astronomy}}} \\
\midrule
Ring\_Moat\_Dome\_Str. & \faImage & 88.6 & \gGood 76.0 & \gHigh 100.0 & \gHigh 100.0 & \gHigh 100.0 & \gHigh 100.0 & \gHigh 100.0 & \gHigh 99.8 & \gMid 67.5 & \gHigh 100.0 & \gMid 73.4 & \gLow 45.7 & \gHigh 93.8 & \gHigh 93.3 & \gGood 89.0 & \gGood 87.0 \\
Ionospheric\_Obs. & \faGlobe & 88.1 & \gGood 76.0 & \gHigh 100.0 & \gHigh 100.0 & \gGood 86.8 & \gGood 82.4 & \gMid 74.3 & \gHigh 100.0 & \gGood 82.8 & \gGood 78.8 & - & \gBad - & \gHigh 100.0 & \gHigh 100.0 & \gGood 87.6 & \gGood 87.4 \\
Martian\_Aeol.\_Land. & \faImage & 86.2 & \gGood 76.0 & \gHigh 100.0 & \gHigh 100.0 & \gGood 75.0 & \gHigh 100.0 & \gHigh 100.0 & \gHigh 100.0 & \gHigh 93.7 & \gHigh 100.0 & \gMid 61.3 & - & \gGood 87.5 & \gGood 86.0 & \gLow 45.3 & \gGood 78.0 \\
SOLAR\_WIND\_MODEL & \faTable & 83.7 & \gGood 80.0 & \gHigh 100.0 & \gHigh 100.0 & \gHigh 98.0 & \gHigh 90.9 & \gHigh 99.9 & \gHigh 100.0 & \gGood 87.5 & \gHigh 99.3 & \gGood 82.0 & \gBad 15.4 & \gHigh 100.0 & \gHigh 97.0 & \gHigh 93.7 & \gBad 24.2 \\
Solar\_Act.\_AI\_Fore. & \faImage,\faTable & 81.8 & \gGood 76.0 & \gHigh 100.0 & \gHigh 100.0 & \gHigh 93.5 & \gGood 88.5 & \gBad 0.0 & \gHigh 100.0 & \gMid 73.3 & \gHigh 100.0 & \gGood 83.4 & \gBad - & \gHigh 100.0 & \gHigh 94.5 & \gMid 61.3 & \gLow 46.7 \\
Optical\_Star\_Image & \faImage & 81.3 & \gGood 76.0 & \gHigh 100.0 & \gHigh 100.0 & \gHigh 100.0 & \gHigh 98.8 & \gHigh 100.0 & \gHigh 100.0 & \gHigh 92.5 & \gMid 51.8 & \gLow 25.7 & - & \gGood 88.5 & \gMid 58.9 & \gGood 75.0 & \gGood 85.2 \\
LAMOST\_DR5 & \faTable & 80.9 & \gHigh 100.0 & \gHigh 100.0 & \gHigh 100.0 & \gGood 81.4 & \gLow 34.4 & \gHigh 100.0 & \gHigh 100.0 & \gHigh 99.1 & \gMid 67.6 & \gMid 52.7 & \gGood 81.0 & \gGood 81.2 & \gHigh 98.0 & \gLow 30.0 & \gMid 57.9 \\
Solar\_WL\_Flares & \faImage,\faTable,\faGlobe & 78.3 & \gGood 76.0 & \gHigh 100.0 & \gHigh 100.0 & \gHigh 100.0 & \gLow 46.5 & \gHigh 100.0 & \gGood 75.5 & \gMid 62.1 & \gHigh 100.0 & \gGood 86.8 & - & \gHigh 100.0 & \gHigh 98.3 & \gLow 40.2 & \gMid 50.0 \\
Solar\_Chrom.\_H$\alpha$ & \faImage,\faGlobe & 78.2 & \gGood 76.0 & \gHigh 100.0 & \gHigh 100.0 & \gHigh 100.0 & \gHigh 99.8 & \gHigh 100.0 & \gGood 77.6 & \gMid 73.1 & \gLow 31.9 & - & - & \gGood 82.7 & \gHigh 94.0 & \gLow 40.3 & \gGood 78.5 \\
Stellar\_Prof.\_L.C. & \faTable & 77.9 & \gGood 76.0 & \gHigh 100.0 & \gHigh 100.0 & \gHigh 100.0 & - & \gHigh 99.6 & \gHigh 100.0 & \gMid 62.1 & \gLow 25.6 & \gBad 13.2 & \gMid 71.6 & \gMid 53.6 & \gMid 71.0 & \gHigh 91.1 & \gHigh 94.3 \\
Footprints\_Solar\_W. & \faImage & 76.9 & \gGood 76.0 & \gHigh 100.0 & \gHigh 100.0 & \gHigh 98.9 & \gGood 77.8 & \gHigh 100.0 & \gGood 80.6 & \gLow 42.6 & \gBad 16.0 & - & \gMid 66.1 & \gMid 68.3 & \gHigh 90.0 & \gGood 83.5 & \gHigh 96.7 \\
Astro.\_Observation & \faGlobe,\faTable & 76.7 & \gGood 76.0 & \gHigh 100.0 & \gHigh 100.0 & \gHigh 95.4 & \gHigh 100.0 & \gBad 24.0 & \gMid 58.9 & \gHigh 100.0 & \gMid 59.4 & - & \gMid 61.6 & \gHigh 100.0 & \gLow 29.8 & \gGood 88.3 & \gMid 68.3 \\
Mag.\_Field\_Config. & \faGlobe & 76.3 & \gGood 80.0 & \gHigh 100.0 & \gHigh 100.0 & \gHigh 99.2 & - & \gHigh 92.3 & \gHigh 95.5 & \gMid 55.9 & \gBad 18.0 & - & - & \gHigh 100.0 & \gMid 70.0 & \gMid 73.3 & \gMid 73.1 \\
origin & \faTable & 73.3 & \gGood 80.0 & \gGood 88.9 & \gHigh 100.0 & \gHigh 100.0 & \gHigh 99.0 & \gHigh 100.0 & \gMid 50.0 & \gMid 61.9 & \gBad 19.0 & \gGood 75.3 & \gBad 17.7 & \gGood 87.5 & \gLow 46.7 & \gMid 66.7 & \gHigh 90.6 \\
Lunar\_Spectroscopy & \faTable & 70.4 & \gGood 76.0 & \gHigh 100.0 & \gHigh 100.0 & \gHigh 94.7 & \gLow 43.9 & \gHigh 100.0 & \gHigh 99.6 & \gBad 19.0 & \gBad 19.0 & - & \gMid 52.9 & \gMid 62.5 & \gHigh 100.0 & \gLow 36.7 & \gHigh 100.0 \\
Circular\_Polarization & \faGlobe & 69.3 & \gGood 76.0 & \gHigh 100.0 & \gHigh 100.0 & \gHigh 94.1 & \gLow 38.8 & \gHigh 94.1 & \gHigh 100.0 & \gBad 21.8 & \gBad 20.6 & - & - & \gMid 68.8 & \gHigh 100.0 & \gHigh 100.0 & \gMid 60.0 \\

\midrule
\rowcolor{green!8} \multicolumn{18}{c}{\textit{\textbf{Biomedicine}}} \\
\midrule
Flysta3D & \faImage & 91.4 & \gGood 85.7 & \gHigh 100.0 & \gGood 79.2 & \gHigh 100.0 & \gHigh 100.0 & \gHigh 100.0 & \gHigh 100.0 & \gHigh 99.6 & \gMid 71.2 & - & \gHigh 100.0 & \gHigh 98.1 & \gHigh 100.0 & \gMid 53.3 & - \\
The-Gonzo-Dataset & \faGlobe,\faTable & 86.8 & \gHigh 100.0 & \gMid 55.6 & \gGood 83.3 & \gHigh 100.0 & \gLow 48.8 & - & \gHigh 100.0 & \gHigh 100.0 & \gGood 78.3 & - & - & \gHigh 100.0 & \gHigh 100.0 & \gGood 82.7 & \gHigh 100.0 \\
Synth.-Skeletal-Mot. & \faImage,\faTable & 81.4 & \gHigh 100.0 & \gLow 44.4 & \gGood 83.3 & \gHigh 100.0 & \gMid 50.0 & \gHigh 96.8 & \gHigh 100.0 & \gGood 87.7 & \gMid 68.0 & \gGood 84.5 & \gHigh 91.0 & \gHigh 92.2 & \gMid 60.0 & \gGood 86.3 & \gGood 87.1 \\
MedMNISTv2 & \faGlobe & 80.2 & \gHigh 100.0 & \gMid 55.6 & \gGood 83.3 & \gHigh 100.0 & \gHigh 100.0 & \gHigh 92.1 & \gHigh 100.0 & \gMid 70.0 & \gGood 75.6 & \gGood 80.3 & - & \gGood 85.0 & \gMid 74.7 & \gBad 6.0 & \gGood 85.2 \\
Genomic-benchmarks & \faFont & 79.9 & \gMid 60.0 & \gMid 66.7 & \gMid 66.7 & \gHigh 90.6 & \gGood 78.1 & \gMid 70.0 & \gHigh 100.0 & \gHigh 94.0 & \gGood 79.9 & \gHigh 98.6 & - & \gMid 62.5 & \gGood 80.0 & \gBad 0.0 & \gHigh 96.1 \\
RNA-seq-human-kidney & \faTable & 79.8 & \gMid 60.0 & \gMid 66.7 & \gMid 62.5 & \gHigh 100.0 & \gHigh 100.0 & \gHigh 100.0 & \gHigh 100.0 & \gHigh 95.3 & \gMid 64.1 & \gHigh 100.0 & \gBad - & \gMid 51.2 & \gHigh 100.0 & \gMid 50.7 & - \\
HQColon\_sampled & \faImage,\faGlobe,\faTable & 77.9 & \gMid 56.0 & \gMid 66.7 & \gMid 70.8 & \gBad 0.0 & \gGood 80.6 & \gBad 0.0 & \gGood 87.5 & \gHigh 98.0 & \gGood 81.1 & \gGood 80.4 & - & \gHigh 100.0 & \gHigh 100.0 & \gGood 76.7 & \gBad 16.7 \\
Insect-Odorant-Bind. & \faProjectDiagram,\faTable & 76.8 & \gHigh 96.0 & \gBad 22.2 & \gGood 79.2 & \gGood 87.5 & \gMid 63.5 & \gHigh 99.5 & \gHigh 94.7 & \gHigh 95.5 & \gMid 64.8 & \gHigh 93.7 & \gLow 25.7 & \gHigh 95.5 & \gGood 86.5 & \gGood 75.0 & \gGood 84.9 \\
RAS-Dataset & \faGlobe & 74.8 & \gHigh 96.0 & \gHigh 100.0 & \gHigh 100.0 & \gGood 75.0 & \gHigh 100.0 & \gHigh 100.0 & \gGood 78.0 & \gGood 86.9 & \gMid 62.5 & \gBad 5.9 & - & \gGood 77.1 & \gHigh 90.0 & \gBad 10.0 & \gMid 64.6 \\
NasalSeg & \faGlobe & 74.5 & \gHigh 96.0 & \gBad 22.2 & \gGood 79.2 & \gHigh 100.0 & \gHigh 90.0 & \gGood 82.3 & \gGood 80.6 & \gHigh 100.0 & \gMid 58.1 & \gHigh 93.1 & - & \gMid 59.7 & \gGood 80.0 & \gLow 30.0 & - \\
ticks\_in\_China & \faTable & 73.9 & \gLow 36.0 & \gMid 55.6 & \gMid 54.2 & \gHigh 93.8 & \gGood 81.9 & \gLow 45.9 & \gHigh 100.0 & \gHigh 90.7 & \gHigh 93.6 & \gBad 0.0 & \gGood 87.1 & \gHigh 100.0 & \gGood 89.5 & \gLow 33.1 & \gGood 81.7 \\
De-novo-transcriptomic & \faFont,\faTable & 72.7 & \gLow 36.0 & \gMid 55.6 & \gMid 58.3 & \gHigh 97.9 & \gHigh 95.5 & \gHigh 93.6 & \gHigh 98.2 & \gGood 79.5 & \gHigh 90.4 & - & \gHigh 91.8 & \gMid 51.8 & \gHigh 100.0 & \gBad 20.0 & - \\
Human-Tongue-Musc. & \faGlobe & 68.9 & \gMid 52.0 & \gBad 22.2 & \gMid 70.8 & \gHigh 96.4 & \gGood 86.3 & \gHigh 100.0 & \gHigh 100.0 & \gMid 73.8 & \gMid 52.7 & \gLow 26.3 & - & \gGood 87.5 & \gGood 87.5 & \gLow 49.3 & - \\
miRTarBase\_MTI.csv & \faTable & 67.8 & \gBad 24.0 & \gBad 0.0 & \gMid 70.0 & \gGood 88.7 & \gHigh 99.5 & \gLow 48.6 & \gHigh 100.0 & \gHigh 95.0 & \gHigh 100.0 & \gLow 33.3 & - & \gMid 65.4 & \gGood 76.6 & \gMid 50.0 & - \\
SUSTech-SYSU & \faImage & 66.3 & \gLow 36.0 & \gMid 66.7 & \gMid 54.2 & \gHigh 100.0 & \gMid 70.2 & \gHigh 100.0 & \gHigh 100.0 & \gGood 89.7 & \gMid 54.5 & \gMid 60.5 & - & \gGood 89.3 & \gMid 59.9 & \gBad 10.0 & \gMid 50.0 \\
\midrule

\bottomrule
\end{tabular*}
\end{table*}

\begin{table*}[t]
\footnotesize
\setlength{\tabcolsep}{4pt}

\caption{Quantitative Evaluation Results — All Domains (Part Part II). Background intensity correlates with score magnitude.
\textbf{$T$}: Governance Trustworthiness ($T_1$: FAIRness, $T_2$: Provenance, $T_3$: Ethics);
\textbf{$Q$}: Data Quality ($Q_1$: Complete., $Q_2$: Acc., $Q_3$: Unique., $Q_4$: Consist.);
\textbf{$A^c$}: AI Compatibility ($A^c_1$: Adapt., $A^c_2$: Scale, $A^c_3$: Bal., $A^c_4$: Feat.Imp.);
\textbf{$A^s$}: Scientific Adaptability ($A^s_1$: Task Gen., $A^s_2$: Scarcity, $A^s_3$: Causal, $A^s_4$: Cond. Cov.).
\textit{Modality}: \faGlobe~Tensor, \faProjectDiagram~Graph, \faTable~Table, \faImage~Image, \faFont~Text.}

\label{tab:leaderboard_appendix_part2}

\centering
\begin{tabular*}{\textwidth}{@{\extracolsep{\fill}} l c c | ccc | cccc | cccc | cccc}
\toprule
\multirow{2}{*}{\textbf{Dataset}} &
\multirow{2}{*}{\textbf{Mod.}} &
\multirow{2}{*}{\textbf{Total}} &
\multicolumn{3}{c|}{\textbf{Gov. Trust. ($T$)}} &
\multicolumn{4}{c|}{\textbf{Data Qual. ($Q$)}} &
\multicolumn{4}{c|}{\textbf{AI Comp. ($A^c$)}} &
\multicolumn{4}{c}{\textbf{Sci. Adapt. ($A^s$)}} \\
\cmidrule(lr){4-6} \cmidrule(lr){7-10} \cmidrule(lr){11-14} \cmidrule(lr){15-18}
 &  &  & $T_1$ & $T_2$ & $T_3$ & $Q_1$ & $Q_2$ & $Q_3$ & $Q_4$ & $A^c_1$ & $A^c_2$ & $A^c_3$ & $A^c_4$ & $A^s_1$ & $A^s_2$ & $A^s_3$ & $A^s_4$ \\
\midrule

\midrule
\rowcolor{cyan!8} \multicolumn{18}{c}{\textit{\textbf{Earth Science}}} \\
\midrule
Fengyun-3 & \faImage & 90.2 & \gGood 76.0 & \gHigh 100.0 & \gHigh 100.0 & \gHigh 93.5 & \gGood 88.0 & \gHigh 100.0 & \gHigh 100.0 & \gMid 73.5 & \gHigh 91.5 & - & - & \gHigh 99.8 & \gHigh 100.0 & \gMid 63.3 & \gHigh 100.0 \\
Space\_Hurr.\_Recog. & \faImage & 86.0 & \gGood 76.0 & \gHigh 100.0 & \gHigh 100.0 & \gHigh 100.0 & \gGood 88.3 & \gLow 45.8 & \gMid 62.9 & \gHigh 90.0 & \gHigh 93.1 & \gHigh 100.0 & - & \gGood 81.5 & \gHigh 93.3 & \gHigh 90.0 & \gHigh 90.4 \\
Turb.\_Heat\_Flux\_2023 & \faGlobe & 85.7 & \gGood 76.0 & \gHigh 100.0 & \gHigh 100.0 & \gGood 78.1 & \gGood 85.2 & \gHigh 100.0 & \gHigh 100.0 & \gHigh 94.0 & \gGood 80.0 & - & \gHigh 100.0 & \gMid 50.4 & \gHigh 95.0 & \gMid 60.0 & \gBad 10.0 \\
LoveDA & \faImage & 85.5 & \gHigh 100.0 & \gMid 55.6 & \gGood 83.3 & \gHigh 100.0 & \gGood 89.3 & \gHigh 100.0 & \gHigh 100.0 & \gGood 87.5 & \gGood 89.8 & \gHigh 94.0 & - & \gGood 77.5 & \gMid 71.7 & \gMid 55.0 & \gHigh 93.9 \\
EM\_Field\_Structure & \faGlobe & 85.2 & \gGood 76.0 & \gHigh 100.0 & \gHigh 100.0 & \gHigh 100.0 & - & - & \gHigh 100.0 & \gHigh 99.9 & \gMid 51.0 & - & - & \gMid 70.8 & \gGood 80.0 & \gGood 84.0 & \gMid 58.5 \\
Mass\_Transport & \faGlobe & 84.5 & \gGood 76.0 & \gHigh 100.0 & \gHigh 100.0 & \gHigh 100.0 & - & - & \gHigh 100.0 & \gHigh 95.5 & \gMid 60.0 & - & - & \gGood 80.5 & \gMid 70.0 & \gMid 53.3 & \gMid 68.4 \\
EuroSAT & \faImage & 78.4 & \gHigh 100.0 & \gMid 55.6 & \gGood 83.3 & \gMid 50.0 & \gHigh 99.0 & \gHigh 100.0 & \gHigh 100.0 & \gGood 88.9 & \gMid 61.5 & \gHigh 99.8 & - & \gGood 81.6 & \gMid 66.6 & \gBad 20.0 & \gGood 85.6 \\
Ionospheric\_Obs. & \faGlobe & 77.7 & \gGood 76.0 & \gHigh 100.0 & \gHigh 100.0 & \gGood 81.8 & \gMid 64.8 & \gHigh 92.6 & \gHigh 96.7 & \gGood 81.0 & \gBad 3.2 & - & - & \gHigh 93.8 & \gHigh 100.0 & \gGood 77.2 & \gHigh 100.0 \\
Airglow\_Met.-Radar\_Sat. & \faImage & 67.7 & \gGood 76.0 & \gHigh 100.0 & \gHigh 100.0 & \gLow 43.6 & \gLow 37.3 & \gHigh 100.0 & \gMid 50.6 & \gMid 53.4 & \gLow 40.1 & - & - & \gHigh 94.6 & \gHigh 92.0 & \gGood 80.8 & \gLow 29.0 \\

\midrule
\rowcolor{orange!8} \multicolumn{18}{c}{\textit{\textbf{Materials \& Chemistry}}} \\
\midrule
FreeSolv & \faTable,\faProjectDiagram & 88.2 & \gGood 85.7 & \gHigh 100.0 & \gGood79.2 & \gHigh 100.0 & \gHigh 100.0 & \gHigh 100.0 & \gHigh 99.8 & \gHigh 100.0 & \gMid 56.2 & \gGood 76.2 & \gHigh 96.8 & \gGood 87.5 & \gHigh 97.5 & \gHigh 100.0 & \gHigh 90.3 \\
BigSolDBv2.0 & \faTable & 88.0 & \gHigh 100.0 & \gMid 55.6 & \gGood 83.3 & \gHigh 99.3 & \gHigh 91.3 & \gHigh 100.0 & \gHigh 100.0 & \gGood 89.2 & \gHigh 100.0 & \gGood 80.1 & \gGood 77.6 & \gMid 68.8 & \gHigh 100.0 & \gHigh 95.0 & \gGood 87.5 \\
xxMD & \faProjectDiagram & 86.3 & \gGood 76.0 & \gHigh 100.0 & \gHigh 100.0 & \gHigh 100.0 & \gHigh 99.3 & \gHigh 100.0 & \gHigh 100.0 & \gHigh 90.6 & \gGood 84.0 & \gMid 71.8 & \gLow 47.6 & \gHigh 100.0 & \gHigh 100.0 & \gLow 48.2 & \gGood 85.4 \\
Raw\_Hyperspec.\_Refl. & \faTable & 85.0 & \gHigh 96.0 & \gHigh 100.0 & \gHigh 95.8 & \gHigh 99.8 & \gHigh 99.4 & \gHigh 100.0 & \gHigh 100.0 & \gGood 81.2 & \gBad 16.0 & \gGood 87.2 & \gMid 73.7 & \gMid 71.2 & \gHigh 100.0 & \gMid 66.7 & \gGood 81.9 \\
1018\_Compounds & \faTable & 85.0 & \gHigh 96.0 & \gHigh 100.0 & \gHigh 95.8 & \gHigh 100.0 & \gMid 71.0 & \gHigh 100.0 & \gHigh 100.0 & \gHigh 99.9 & \gMid 68.1 & \gHigh 99.4 & - & \gMid 62.5 & \gHigh 90.0 & \gLow 40.0 & \gMid 72.0 \\
CO2\_Electrocat.\_Red. & \faTable & 81.9 & \gHigh 96.0 & \gHigh 100.0 & \gHigh 100.0 & \gHigh 98.8 & \gHigh 97.5 & \gHigh 100.0 & \gHigh 97.3 & \gHigh 100.0 & \gGood 88.1 & \gLow 38.7 & - & \gMid 68.4 & \gGood 81.6 & \gMid 57.8 & \gGood 75.0 \\
ML-Driven\_Elast.\_Pred. & \faTable & 78.7 & \gGood 76.0 & \gHigh 100.0 & \gHigh 100.0 & \gHigh 100.0 & \gHigh 91.8 & \gHigh 100.0 & \gHigh 100.0 & \gHigh 98.0 & \gGood 87.5 & \gMid 70.0 & \gMid 55.3 & \gMid 62.5 & \gMid 50.4 & \gLow 44.0 & \gLow 40.2 \\
Hydrogen\_Combustion & \faGlobe & 73.2 & \gLow 36.0 & \gMid 55.6 & \gMid 54.2 & \gHigh 100.0 & - & \gHigh 90.0 & \gHigh 100.0 & \gGood 80.5 & \gHigh 100.0 & \gBad 2.0 & - & \gHigh 96.7 & \gHigh 93.0 & \gMid 67.0 & \gGood 89.4 \\
USPTO\_LLM & \faProjectDiagram & 73.2 & \gHigh 100.0 & \gMid 55.6 & \gGood 83.3 & \gHigh 100.0 & \gGood 83.6 & \gHigh 100.0 & \gHigh 100.0 & \gGood 80.0 & \gHigh 100.0 & \gBad 13.7 & - & \gMid 62.5 & \gMid 67.5 & \gLow 49.6 & \gMid 71.5 \\
Open-Frmwk.\_AlPO\_Str. & \faProjectDiagram & 71.9 & \gLow 36.0 & \gMid 55.6 & \gMid 54.2 & \gHigh 96.4 & \gGood 89.4 & \gHigh 100.0 & \gHigh 99.6 & \gHigh 97.5 & \gGood 87.4 & \gLow 30.0 & \gLow 46.9 & \gMid 70.8 & \gHigh 97.5 & \gMid 71.2 & \gMid 68.9 \\

\midrule
\rowcolor{red!8} \multicolumn{18}{c}{\textit{\textbf{Physics
\& Engineering}}} \\
\midrule
Multiharm.\_Feedback & \faTable & 93.2 & \gHigh 100.0 & \gHigh 100.0 & \gHigh 100.0 & \gHigh 100.0 & \gHigh 100.0 & \gHigh 100.0 & \gHigh 100.0 & \gHigh 100.0 & \gGood 83.9 & - & \gGood 82.4 & \gMid 66.7 & \gGood 85.0 & \gGood 85.3 & \gHigh 100.0 \\
CdS\_Nanowire\_SHG & \faTable & 88.9 & \gHigh 92.0 & \gHigh 100.0 & \gHigh 100.0 & \gHigh 100.0 & \gMid 55.6 & \gHigh 100.0 & \gHigh 100.0 & \gHigh 96.0 & \gLow 49.4 & - & \gGood 86.6 & \gGood 87.5 & \gHigh 90.0 & \gHigh 90.0 & \gHigh 100.0 \\
Dusty\_Plasma\_Ratchet & \faTable & 86.8 & \gHigh 96.0 & \gHigh 100.0 & \gHigh 100.0 & \gHigh 100.0 & - & \gHigh 100.0 & \gHigh 92.9 & \gGood 86.2 & \gMid 70.3 & \gMid 55.5 & \gHigh 94.5 & \gHigh 93.8 & \gLow 25.0 & \gGood 82.9 & \gGood 76.7 \\
Tokamak\_Dust\_Dynamics & \faTable & 85.1 & \gGood 76.0 & \gHigh 100.0 & \gHigh 100.0 & \gHigh 100.0 & \gHigh 100.0 & \gHigh 100.0 & \gHigh 100.0 & \gMid 65.1 & \gMid 55.0 & - & \gHigh 100.0 & \gMid 70.0 & \gHigh 100.0 & \gGood 83.3 & \gLow 47.0 \\
Coherent\_Light\_Int. & \faTable & 85.0 & \gGood 76.0 & \gHigh 100.0 & \gHigh 100.0 & \gHigh 100.0 & \gGood 80.0 & \gHigh 100.0 & \gHigh 100.0 & \gHigh 100.0 & \gMid 65.0 & - & - & \gGood 81.2 & \gMid 70.0 & \gGood 76.5 & \gMid 55.0 \\
Laser\_Chirp\_EUV & \faImage & 82.2 & \gGood 76.0 & \gHigh 100.0 & \gHigh 100.0 & \gMid 66.7 & \gGood 86.7 & \gHigh 100.0 & \gMid 63.7 & \gMid 67.7 & \gMid 60.9 & \gGood 86.7 & - & \gHigh 100.0 & \gGood 80.0 & \gGood 82.9 & \gHigh 98.9 \\
FPD\_Rotor\_Aeroelastic & \faGlobe & 80.3 & \gGood 80.0 & \gHigh 100.0 & \gHigh 100.0 & \gHigh 99.3 & \gHigh 93.0 & \gHigh 97.6 & \gHigh 99.1 & \gMid 66.2 & \gLow 30.6 & \gBad 13.8 & \gLow 49.2 & \gHigh 100.0 & \gHigh 100.0 & \gHigh 96.7 & \gMid 66.8 \\
Quantum\_Metric\_Tensor & \faTable,\faImage & 80.1 & \gHigh 96.0 & \gMid 66.7 & \gGood 75.0 & \gMid 50.1 & \gHigh 100.0 & \gHigh 99.3 & \gHigh 100.0 & \gGood 78.2 & \gMid 65.0 & - & \gMid 73.2 & \gHigh 100.0 & \gHigh 100.0 & \gGood 76.7 & \gMid 50.0 \\
fMeta-TPC & \faTable & 79.9 & \gGood 80.0 & \gHigh 100.0 & \gHigh 100.0 & \gHigh 100.0 & \gGood 77.3 & \gBad 17.1 & \gMid 72.9 & \gHigh 96.7 & \gGood 81.3 & - & \gGood 75.1 & \gMid 62.5 & \gGood 80.0 & \gGood 76.4 & \gGood 82.0 \\
HIGGS & \faTable & 78.0 & \gMid 52.0 & \gBad 0.0 & \gGood 75.0 & \gHigh 100.0 & \gGood 86.2 & \gBad 0.0 & \gHigh 100.0 & \gHigh 100.0 & \gHigh 100.0 & \gHigh 99.9 & \gHigh 93.8 & \gLow 48.2 & \gMid 70.0 & \gGood 85.0 & \gGood 84.0 \\
Proton\_Range & \faTable & 76.9 & \gHigh 100.0 & \gHigh 100.0 & \gHigh 100.0 & \gLow 43.5 & \gGood 81.7 & \gMid 68.1 & \gHigh 97.6 & \gHigh 93.4 & \gBad 8.7 & \gMid 67.0 & \gBad 7.3 & \gHigh 100.0 & \gGood 82.0 & \gGood 81.7 & \gHigh 99.5 \\
Quantum\_Gas\_Light & \faImage & 71.9 & \gHigh 96.0 & \gBad 22.2 & \gGood 79.2 & \gHigh 94.4 & \gGood 89.6 & \gHigh 99.7 & \gGood 86.9 & \gBad 22.7 & \gMid 50.2 & - & - & \gHigh 93.0 & \gHigh 100.0 & \gGood 85.0 & \gHigh 92.4 \\
PSF\_Photolithography & \faImage & 68.0 & \gGood 76.0 & \gHigh 100.0 & \gHigh 95.8 & \gHigh 100.0 & \gHigh 100.0 & \gHigh 100.0 & \gHigh 100.0 & \gLow 40.0 & \gBad 12.7 & \gMid 70.0 & - & \gLow 47.0 & \gMid 55.0 & \gMid 54.7 & \gLow 35.0 \\

\midrule
\rowcolor{yellow!8} \multicolumn{18}{c}{\textit{\textbf{Socio-economics}}} \\
\midrule
Dig.\_Econ.\_Green\_Dev. & \faTable & 95.3 & \gHigh 100.0 & \gHigh 100.0 & \gHigh 100.0 & \gHigh 100.0 & - & \gHigh 100.0 & \gHigh 100.0 & \gGood 82.5 & \gHigh 93.1 & \gBad - & \gGood 85.3 & \gHigh 100.0 & \gHigh 100.0 & \gHigh 96.7 & \gGood 81.0 \\
Env.\_Reg.\_Green\_Inn. & \faTable & 91.0 & \gHigh 100.0 & \gHigh 100.0 & \gHigh 100.0 & \gHigh 100.0 & - & \gHigh 100.0 & \gHigh 100.0 & \gHigh 100.0 & \gGood 87.5 & - & \gMid 66.0 & \gHigh 100.0 & \gMid 56.8 & \gHigh 95.3 & \gMid 57.6 \\
China\_Cross\_Bord.\_Ecom. & \faTable & 89.2 & \gHigh 100.0 & \gHigh 100.0 & \gHigh 100.0 & \gMid 74.1 & \gHigh 100.0 & \gHigh 100.0 & \gHigh 100.0 & \gHigh 99.8 & \gHigh 96.4 & \gLow 49.4 & \gMid 72.6 & \gHigh 100.0 & \gHigh 99.7 & \gHigh 100.0 & \gGood 85.0 \\
China\_282\_Green\_Econ. & \faTable & 88.6 & \gGood 76.0 & \gHigh 100.0 & \gHigh 100.0 & \gHigh 100.0 & - & \gHigh 100.0 & \gHigh 100.0 & \gGood 75.0 & \gGood 88.6 & \gMid 71.3 & \gGood 88.1 & \gHigh 100.0 & \gGood 77.6 & \gGood 75.0 & \gGood 87.3 \\
Mgmt.\_Tone\_Stock\_Pr. & \faTable & 87.5 & \gGood 80.0 & \gGood 88.9 & \gHigh 100.0 & \gHigh 100.0 & \gMid 50.0 & \gHigh 100.0 & \gHigh 100.0 & \gGood 89.1 & \gMid 73.4 & - & \gGood 87.4 & \gHigh 100.0 & \gHigh 95.0 & \gGood 75.8 & \gGood 86.7 \\
W.\_China\_Farmer\_Surv. & \faTable & 86.7 & \gHigh 100.0 & \gHigh 100.0 & \gHigh 100.0 & \gHigh 98.0 & - & \gHigh 100.0 & \gHigh 99.9 & \gHigh 90.2 & \gLow 45.6 & \gHigh 100.0 & \gHigh 98.6 & \gGood 76.7 & \gHigh 90.0 & \gLow 37.0 & - \\
Dig.\_Incl.\_Rural\_Cons. & \faTable & 86.0 & \gHigh 100.0 & \gHigh 100.0 & \gHigh 100.0 & \gHigh 97.4 & - & \gHigh 100.0 & \gHigh 100.0 & \gGood 81.4 & \gMid 73.8 & \gMid 63.8 & \gMid 57.7 & \gGood 87.5 & \gHigh 95.3 & \gMid 50.0 & - \\
Serv.\_Ind.\_Econ.\_Res. & \faTable & 85.7 & \gHigh 100.0 & \gHigh 100.0 & \gHigh 100.0 & \gHigh 100.0 & - & \gHigh 100.0 & \gHigh 100.0 & \gMid 73.9 & \gGood 79.6 & \gLow 49.6 & \gGood 77.4 & \gHigh 100.0 & \gMid 57.5 & \gMid 60.0 & \gMid 70.0 \\
Google\_COVID\_Mobility & \faTable & 85.1 & - & - & - & \gGood 76.0 & \gHigh 95.0 & \gHigh 100.0 & \gHigh 100.0 & \gHigh 98.6 & \gMid 70.7 & - & \gMid 68.7 & \gGood 76.7 & \gHigh 90.0 & \gGood 83.7 & \gGood 82.8 \\
China\_Green\_Low\_C. & \faTable & 84.6 & \gGood 80.0 & \gMid 66.7 & \gHigh 100.0 & \gHigh 100.0 & - & \gHigh 100.0 & \gHigh 100.0 & \gHigh 95.7 & \gLow 40.7 & - & \gHigh 100.0 & \gHigh 92.4 & \gMid 66.0 & \gMid 65.0 & - \\
COMPAS\_Analysis & \faTable & 82.3 & - & - & - & \gHigh 98.7 & \gGood 78.6 & \gGood 76.0 & \gHigh 100.0 & \gHigh 95.1 & \gHigh 91.1 & \gLow 42.9 & \gLow 48.3 & \gHigh 100.0 & \gHigh 100.0 & \gMid 60.4 & \gHigh 97.4 \\
Cross\_Ling.\_Text\_Cls. & \faFont & 81.6 & \gGood 80.0 & \gHigh 100.0 & \gHigh 100.0 & \gHigh 100.0 & \gHigh 97.4 & \gHigh 93.5 & - & \gHigh 99.0 & \gLow 44.0 & \gHigh 100.0 & - & \gGood 89.8 & \gLow 29.8 & \gMid 71.2 & \gBad 15.4 \\
Childless\_Elder.\_Surv. & \faTable & 77.8 & \gHigh 96.0 & \gHigh 100.0 & \gHigh 95.8 & \gHigh 100.0 & \gMid 50.0 & \gHigh 100.0 & \gHigh 100.0 & \gHigh 92.5 & \gLow 44.1 & \gLow 43.2 & \gMid 59.7 & \gGood 86.2 & \gHigh 95.0 & \gBad 15.0 & \gMid 70.0 \\
Econ\_Sent.\_Dict. & \faFont & 77.1 & \gGood 80.0 & \gHigh 100.0 & \gHigh 100.0 & \gMid 50.0 & \gMid 50.0 & \gHigh 99.3 & \gHigh 100.0 & \gGood 75.2 & \gHigh 100.0 & \gBad - & - & \gLow 25.0 & \gMid 64.3 & \gBad 22.0 & \gHigh 100.0 \\
Yangtze\_Delta\_Vital. & \faTable & 76.7 & \gGood 80.0 & \gHigh 100.0 & \gHigh 100.0 & \gHigh 100.0 & - & \gHigh 100.0 & \gHigh 100.0 & \gGood 84.2 & \gLow 35.0 & - & \gMid 69.7 & \gGood 77.0 & \gMid 64.0 & \gBad 10.0 & - \\
Commer.\_Pension\_Ins. & \faTable & 71.8 & \gGood 80.0 & \gHigh 100.0 & \gHigh 100.0 & \gBad 17.6 & \gHigh 93.0 & \gBad 13.5 & \gHigh 100.0 & \gHigh 94.7 & \gHigh 98.1 & \gMid 69.0 & \gMid 66.7 & \gGood 86.7 & \gMid 60.0 & \gBad 17.8 & - \\
China\_Reg.\_Inequal. & \faTable & 51.9 & \gHigh 100.0 & \gHigh 100.0 & \gHigh 100.0 & \gLow 37.5 & - & \gGood 77.0 & \gHigh 100.0 & \gBad 23.5 & \gBad 0.3 & - & \gBad 0.0 & \gBad 19.0 & \gLow 45.8 & \gBad 7.5 & - \\

\bottomrule
\end{tabular*}
\end{table*}

%% file: tables/appendix/evidence.tex
\begin{table*}[t]
\centering
\caption{Evidence table of datasets, dimensions, scores, and supporting quotations (Part I).}
\label{tab:evidence_dataset_part1}

\setlength{\tabcolsep}{4pt}
\renewcommand{\arraystretch}{1.18}

\begin{tabularx}{\textwidth}{p{2.6cm}|p{2.8cm} c p{5.0cm} X}
\toprule
\textbf{Dataset} & \textbf{Dimension} & \textbf{Score} & \textbf{Article} & \textbf{Evidence} \\
\midrule

\multirow{2}{=}{\texttt{Circular\textunderscore Polarization}} &
Data Scarcity & 100.0 &
Linear to circular conversion in the polarized radio emission of a magnetar, \textit{Nature Astronomy}, 2024 &
``The RM evolutions of these objects are probably the result of slow changes in the particle density and magnetic fields of their local environments. More recently, large RM variations and significant levels of circular polarization have been detected in at least three other repeating FRB sources.'' \\
\cmidrule(l){2-5}
& Data Scarcity & 100.0 &
A persistently active fast radio burst source embedded in an expanding supernova remnant, \textit{Science Bulletin}, 2025 &
``The complexity of its environment is also reflected in the detection of rare circular polarization and a highly stochastic, Brownian motion--like bursting behavior in the time--energy domain.'' \\
\midrule



\multirow{2}{*}{\texttt{BigSolDBv2}} &
Class Balance & 80.1 &
\multirow{2}{=}{BigSolDB 2.0, dataset of solubility values for organic compounds in different solvents at various temperatures, \textit{Scientific Data}, 2025} &
``In this study, we present a dataset containing 103944 experimental solubility values within a temperature range from 243 to 425 K for 1448 organic compounds measured in 213 individual solvents extracted from 1595 peer-reviewed articles.'' \\
\cmidrule(l){2-3}\cmidrule(l){5-5}
& AI Model Adaptability & 89.2 & &
``Moreover, this dataset paves the way to train machine learning models for solubility prediction in a wide variety of organic solvents which can be extremely useful in the design of many chemical and technological processes.'' \\
\midrule
\multirow{2}{*}{\texttt{FreeSolv}} &
Completeness & 100.0 &
\multirow{2}{=}{{FreeSolv original paper / as cited in your manuscript}} &
``The starting point in constructing the FreeSolv database was to pull together all of the lead author’s previous work calculating hydration free energies in explicit solvent. This included calculated values, experimental values, and structures and input files from several previous studies.’’ \\
\cmidrule(l){2-3}\cmidrule(l){5-5}
& Accuracy & 100.0 & &
``we had retained not only calculated and experimental hydration free energies and original coordinate files (.mol2 format) containing geometries and partial charges, but also input files in the form of GROMACS topology and coordinate files.'' \\

\bottomrule
\end{tabularx}
\end{table*}

\begin{table*}[t]
\centering
\caption{Evidence table of datasets, dimensions, scores, and supporting quotations (Part II).}
\label{tab:evidence_dataset_part2}

\setlength{\tabcolsep}{4pt}
\renewcommand{\arraystretch}{1.18}

\begin{tabularx}{\textwidth}{p{2.6cm}|p{2.8cm} c p{5.0cm} X}
\toprule
\textbf{Dataset} & \textbf{Dimension} & \textbf{Score} & \textbf{Article} & \textbf{Evidence} \\
\midrule

\multirow{3}{*}{\texttt{EuroSAT}} &
Class Balance & 99.8 &
\multirow{2}{=}{EuroSAT: A Novel Dataset and Deep Learning Benchmark for Land Use and Land Cover Classification, \textit{IEEE JSTARS}, 2019} &
``We present a novel dataset, based on these images that covers 13 spectral bands and is comprised of ten classes with a total of 27 000 labeled and geo-referenced images.'' \\
\cmidrule(l){2-3}\cmidrule(l){5-5}
& Task Generalizability & 81.6 & &
``The resulting classification system opens a gate toward a number of earth observation applications...'' \\
\cmidrule(l){2-5}
& AI Model Adaptability & 88.9 &
Toward Explainable AI in Satellite Imagery: A ResNet-50-Based Study on EuroSAT Classification, \textit{IEEE Access}, 2025 &
``Leveraging a pre-trained ResNet-50 model on the EuroSAT dataset...'' \\
\midrule

\multirow{5}{*}{\texttt{MedMNISTv2}} &
Completeness & 100.0 &
\multirow{4}{=}{MedMNIST v2 - A large-scale lightweight benchmark for 2D and 3D biomedical image classification, \textit{Scientific Data}, 2023} &
``we provide an official train-validation-test split for each subset... to avoid data leakage'' \\
\cmidrule(l){2-3}\cmidrule(l){5-5}
& Accuracy & 100.0 & &
``Three neuroscience experts segment a pyramidal neuron within the whole volume and proofread all the synapses'' \\
\cmidrule(l){2-3}\cmidrule(l){5-5}
& AI Model Adaptability & 70.0 & &
``For MedMNIST2D, we first implement ResNets10... as baseline methods'' \\
\cmidrule(l){2-3}\cmidrule(l){5-5}
& Task Generalizability & 85.0 & &
``MedMNIST has been particularly useful for machine learning and computer vision research...'' \\
\cmidrule(l){2-3}\cmidrule(l){5-5}
& Consistency & 80.3 & &
``All images are pre-processed into a MNIST-like format...'' \\


\midrule
\multirow{2}{*}{\texttt{ticks\_in\_China}} &
Accuracy & 81.9 &
\multirow{2}{=}{A dataset of distribution and diversity of ticks in China, \textit{Scientific Data}, 2019} &
``After the data were entered, a second person checked the dataset thoroughly to avoid errors and duplications'' \\
\cmidrule(l){2-3}\cmidrule(l){5-5}
& Task Generalizability & 100.0 & &
``The dataset described here can be used to investigate the spatio-temporal dynamics of tick distribution at multiple scales'' \\
\midrule
\multirow{3}{*}{\texttt{miRTarBase\_MTI}} &
Accuracy & 99.5 &
\multirow{2}{=}{miRTarBase update 2022: an informative resource for experimentally validated miRNA--target interactions, \textit{Nucleic Acids Research}, 2022} &
``The database has accumulated >2 200 449 verified MTIs from 13 389 manually curated articles and CLIP-seq data.'' \\
\cmidrule(l){2-3}\cmidrule(l){5-5}
& Task Generalizability & 65.4 & &
``miRTarBase can be applied to miRNA-related disease treatment and drug development.'' \\

\bottomrule
\end{tabularx}
\end{table*}

\begin{table*}[t]
\centering
\caption{Evidence table of datasets, dimensions, scores, and supporting quotations (Part III).}
\label{tab:evidence_dataset_part3}

\setlength{\tabcolsep}{4pt}
\renewcommand{\arraystretch}{1.18}

\begin{tabularx}{\textwidth}{p{2.6cm}|p{2.8cm} c p{5.0cm} X}
\toprule
\textbf{Dataset} & \textbf{Dimension} & \textbf{Score} & \textbf{Article} & \textbf{Evidence} \\
\midrule

\multirow{3}{*}{\texttt{SUSTech-SYSU}} &
Accuracy & 70.2 &
\multirow{3}{=}{The SUSTech-SYSU dataset for automatically segmenting and classifying corneal ulcers, \textit{Scientific Data}, 2020} &
``The quality of all the 712 fluorescein staining images... examined...'' \\
\cmidrule(l){2-3}\cmidrule(l){5-5}
& AI Model Adaptability & 89.7 & &
``this dataset is not only useful for developing and evaluating automated...'' \\
\cmidrule(l){2-3}\cmidrule(l){5-5}
& Task Generalizability & 89.3 & &
``useful for developing and evaluating automated... but also beneficial for...'' \\
\midrule

\multirow{3}{*}{\texttt{COMPAS\_Analysis}} &
Completeness & 98.7 &
\multirow{3}{=}{The accuracy, fairness, and limits of predicting recidivism, \textit{Science Advances}, 2018} &
``This database of 7214 defendants contains...'' \\
\cmidrule(l){2-3}\cmidrule(l){5-5}
& AI Model Adaptability & 95.1 & &
``despite COMPAS's collection of 137 features, the same accuracy can be achieved...'' \\
\cmidrule(l){2-3}\cmidrule(l){5-5}
& Task Generalizability & 100.0 & &
``Data-driven, decision-making technologies used in the justice system... are biased...'' \\
\midrule

\multirow{3}{*}{\texttt{HIGGS}} &
Completeness & 100.0 &
\multirow{2}{=}{Searching for exotic particles in high-energy physics with deep learning, \textit{Nature Communications}, 2014} &
``The data was produced using Monte Carlo simulations...'' \\
\cmidrule(l){2-3}\cmidrule(l){5-5}
& AI Model Adaptability & 100.0 & &
``Figure 7 and Table 1 show the signal efficiency...'' \\
\cmidrule(l){2-5}
& AI Model Adaptability & 100.0 &
XGBoost: A Scalable Tree Boosting System, \textit{KDD}, 2016 &
``we evaluate the performance of XGBoost... on Higgs-1M data...'' \\



\midrule

\multirow{2}{=}{\texttt{Quantum\textunderscore Metric\textunderscore Tensor}} &
Accuracy & 100.0 &
\multirow{2}{=}{Direct probing the quantum geometric tensor for bosonic collective excitations, \textit{arXiv:2601.13963}, 2026} &
``...extracted the quantum metric components... measuring the corresponding pseudospins.'' \\
\cmidrule(l){2-3}\cmidrule(l){5-5}
& Task Generalizability & 100.0 & &
``While the QMT has been experimentally accessed... these measurements have so far been confined...'' \\

\bottomrule
\end{tabularx}
\end{table*}